\documentclass[10pt,twocolumn,twoside]{IEEEtran}
\usepackage{dsfont}
\usepackage{mathtools}
\usepackage{url}
\usepackage{algorithm}
\usepackage{algorithmic}
\usepackage{cite}
\usepackage{xcolor}
\newcommand{\BlackBox}{\rule{1.5ex}{1.5ex}}  
\newenvironment{proof}{\par\noindent{\bf Proof\
}}{\hfill\BlackBox\\[2mm]}

\newtheorem{theorem}{Theorem}

\usepackage{amssymb}
\usepackage{xr}
\usepackage{graphicx}
\usepackage{subfigure}
\usepackage{multirow}
\usepackage{psfrag}

\ifCLASSINFOpdf
\else
\fi
\title{Confidence-Constrained Maximum Entropy Framework for Learning from Multi-Instance Data}
\author{Behrouz~Behmardi,~Forrest~Briggs,~Xiaoli Z. Fern, and~Raviv~Raich\thanks{This work was partially supported by the National Science Foundation grant CCF-1254218.}\\
 School of EECS, Oregon State University, Corvallis, OR, 97331-5501\\
\{behmardb,briggsf,xfern,raich\}@eecs.oregonstate.edu}

\begin{document}

\maketitle

\begin{abstract}

Multi-instance data, in which each object (bag) contains a collection of instances, are widespread in machine learning, computer vision, bioinformatics, signal processing, and social sciences. We present a maximum entropy (ME) framework for learning from multi-instance data. In this approach each bag is represented as a distribution using the principle of ME. We introduce the concept of confidence-constrained ME (CME) to simultaneously learn the structure of distribution space and infer each distribution. The shared structure underlying each density is used to learn from instances inside each bag. The proposed
CME is free of tuning parameters. We devise a fast optimization algorithm capable of handling large scale multi-instance data. In the experimental section, we evaluate the performance of the proposed approach in terms of exact rank recovery in the space of distributions and compare it with the regularized ME approach. Moreover, we compare the performance of CME with Multi-Instance Learning (MIL) state-of-the-art algorithms and show a comparable performance in terms of accuracy with reduced computational complexity.
\end{abstract}


%
\IEEEpeerreviewmaketitle

\section{Introduction}
\begin{figure}[ht]
\centering
\subfigure [] {
\resizebox{5cm}{!}{
\includegraphics{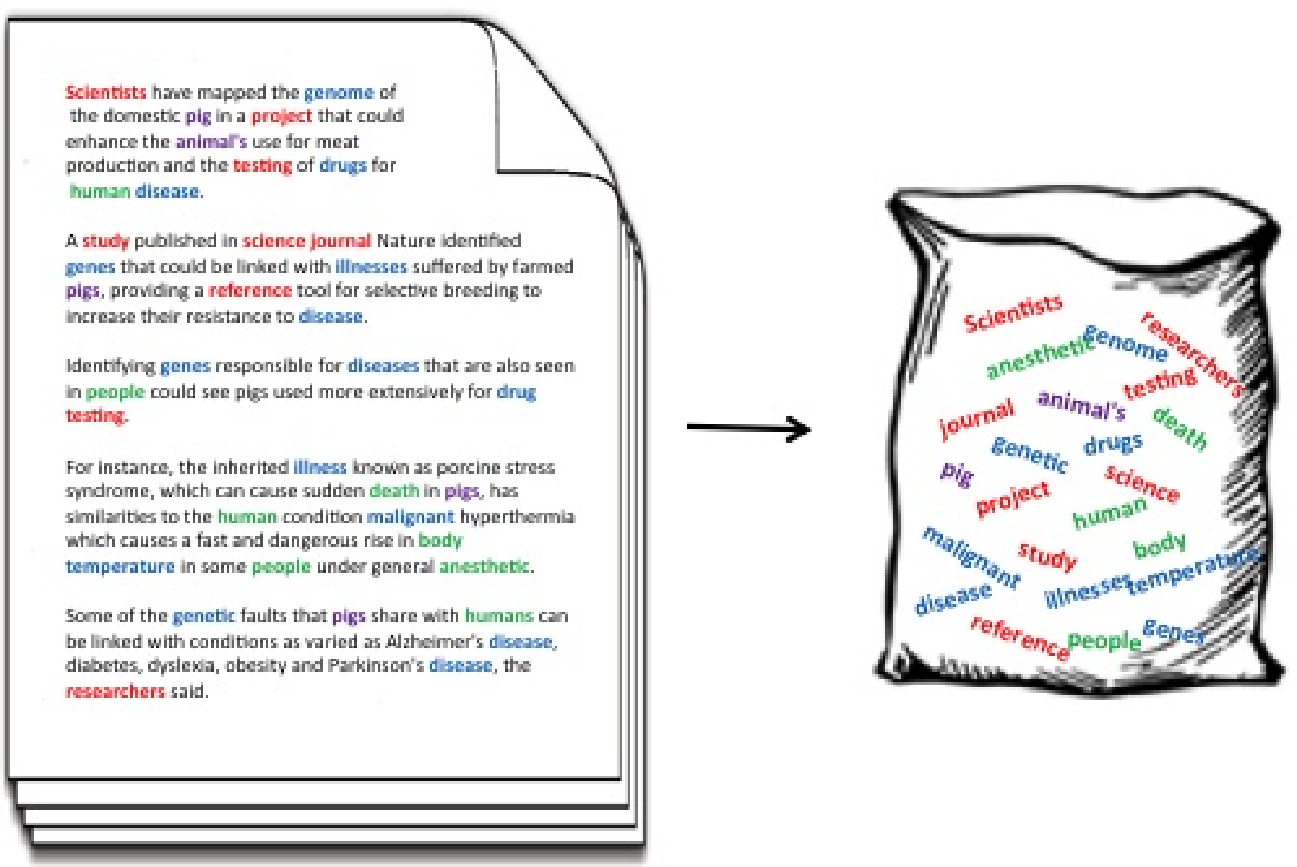}} \label{fig:text}
}
\subfigure [] {
\resizebox{5cm}{!}{
\includegraphics{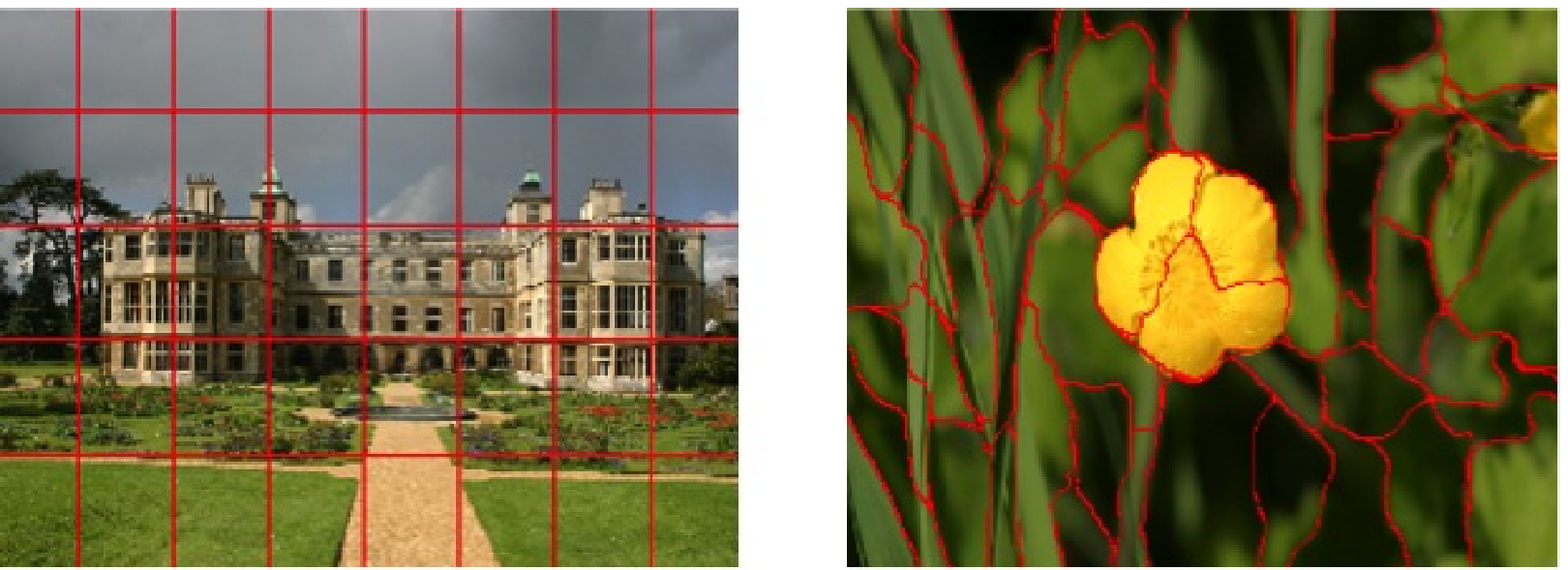}} \label{fig:image}
}
\caption{Multi-instance representation for $(a)$ text documents and $(b)$ images.}\label{fig:mi:example}
\end{figure}

In the multi-instance data representation, objects are viewed as bags of instances.  For example, a document can be viewed as a bag of words, an image can be viewed as bag of segments, and a webpage can be viewed as a bag of links (see Fig.~\ref{fig:mi:example}).
Multi-instance data representation has been used in many areas in machine learning and signal processing, e.g., drug activity detection \cite{dietterich1997solving}, multi-task learning\cite{ni2008multi}, text classification \cite{andrews2002support}, music analysis \cite{qi2007music}, object detection in image \cite{viola2006multiple}, and content-based image categorization \cite{zhang2002content}. Machine learning algorithms are described as either \emph{supervised} or \emph{unsupervised}. Multi-Instance Learning (MIL) refers to the prediction or supervised learning problem \cite{Wang2000,andrews2002support} in which the main goal is to predict the label of an unseen bag, given the label information of the training bags. On the other hand, unsupervised learning (also referred to as grouped data modeling \cite{blei2003latent}) can be applied to unlabeled multi-instance data with the goal of uncovering an underlying structure and a representation for each bag in a collection of multiple instance bags. In supervised MIL, each bag is associated with a class label and the goal is to predict the label for an unseen bag given all the instances inside the bag. Due to the ambiguity of the label information related to instances, supervised MIL is a challenging task. Since the introduction of the MIL approach in machine learning and signal processing,  numerous algorithms have been proposed either by extending traditional algorithms to MIL, e.g., citation kNN \cite{Wang2000}, MI-SVM and mi-SVM \cite{andrews2002support}, neural network MIL \cite{ramon2000multi}, or devising a new algorithm specifically for MIL, e.g., axis-parallel rectangles (APR) \cite{dietterich1997solving}, diverse density (DD) \cite{maron1998framework}, EM-DD \cite{zhang2001dd}, and MIBoosting \cite{xu2004logistic}. MIL has been studied in an unsupervised setting in \cite{zhang2009multi}.

Many of the aforementioned algorithms may compute bag-level similarly metrics (e.g., Haussdoff distance or Mahalanobis distance) based on instance-level similarity \cite{xu2011multi}. Instance-level metrics can become computationally expensive. Computation of  pairwise similarities between of pairs of instance from two given bags involves a computational complexity that increases quadratically in the number of instances in each bag. Moreover, instance level metrics may not reflect the structure similarity defined at the bag level and it is difficult to identify the characteristics of each bag using instance-level similarities \cite{gärtner2002multi}.
Some kernel approaches have been proposed to measure the similarity at the bag-level \cite{gärtner2002multi}. This approach enables the use of kernel based methods for single instance representation to be extened to the bag-level. However, the kernel computation is quadratic in the number of instances per bag.  The problem of computational complexity associated with instance-level metrics has been alleviated by representing each bag with few samples in a very high dimension, e.g., single-blob-with-neighbors (SBN) representation for each image \cite{maron1998multiple}.

\begin{figure*}[ht]
\centering
\includegraphics[scale = 0.8]{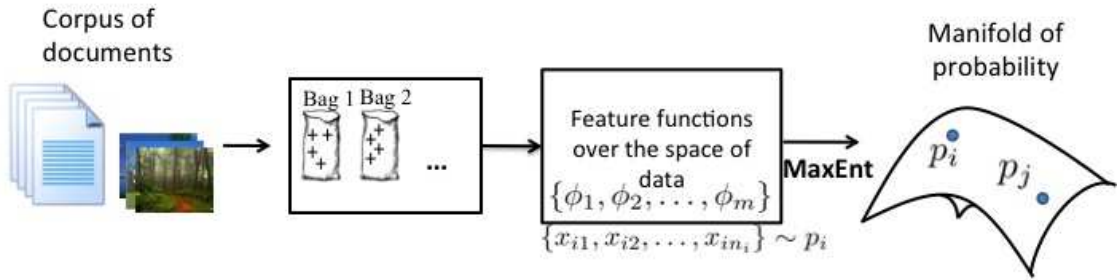}
\caption{Maximum entropy framework used to fit a density to instances (e.g., words) extracted from documents. }\label{fig:MaxEntExample}
\end{figure*}

In this paper, we consider the problem of associating each bag with a probability distribution obtained by the principle of maximum entropy. Assuming that each instance in a bag is generated $i.i.d.$ from an unknown distribution, we fit to each bag a distribution while maintaining a common shared structure among bags (see Fig.~\ref{fig:MaxEntExample}).
This approach has several advantages over existing approaches. First, the problem can be solved in a convex framework. Second, it maps each bag of instances into a point in the probability distribution space providing a summarized representation to the data. In this framework, each bag is parametrized by a vector that carries all the information about the instances inside each bag.
Third, a meaningful metric can be defined over the space of distributions to measure the similarity among bags. Moreover, the computational complexity significantly drops from quadratic to linear in the number of instances inside each bag. The joint density estimation framework with regularization facilitates dimension reduction in the distribution space and introduces a sparse representation of bases which span the space of distributions. Using sparse representations of basis functions to learn the space of distributions in a non-parametric framework has been studied in \cite{paisley2010active}.

Our contributions in this paper are: $1)$ we introduce a new framework for learning from multi-instance data using the principle of maximum entropy, $2)$ a metric defined over the space of the distributions is introduced to measure the similarities among bags in multi-instance data, $3)$ we propose the  confidence-constrained maximum entropy (CME) method to learn the space of distributions jointly, $4)$ an accelerated proximal gradient approach is proposed to solve the resulting convex optimization problem, $5)$ the performance of the proposed approach is evaluated in terms of exact rank recovery in the space of distributions and compared with regularized ME, and $6)$ we examine the classification accuracy of CME on four real world datasets and compared the results with MIL state-of-the-art algorithms.

\section{Problem statement}\label{se:2}
We are given $N$ bags $X_1,X_2, \ldots,X_N$. The $i$th bag $X_i$ given by $X_i=\{x_{i1}, x_{i2},\ldots, x_{in_i}\}$, is a set of $n_i$ feature vectors. Each feature vector $x_{ij}\in\mathcal{X}$ and $\mathcal{X}\subset \mathds{R}^d$. In addition, we assume that the bags are statistically independent and that instances within the same bag are independent and identically distributed ($i.i.d.$). More specifically, for the $i$th bag we assume that $x_{ij}$ for $j=1,2,\ldots,n_i$ are drawn $i.i.d.$ from a probability density function $p_i(x)$. We consider the problem of unsupervised learning of distribution for each bag using the maximum entropy framework. The goals are to $1)$ provide a latent representation for each bag $X_i$ using a generative model $p_{i}(x)$ obtained by maximum entropy and $2)$ provide a joint probability framework with a regularization which takes into account the model complexity and limited number of samples.


\section{Maximum entropy framework for learning from multi-instance data}\label{se:3}
We consider the maximum entropy framework for modeling multi-instance data by associating multi-instance bags with  probability distributions. We
are interested in a framework that will allow convenient incorporation of structure (e.g., geometric, low-dimension) in the distribution space. The problem of density estimation can be define as follows. Given an $i.i.d.$ set of samples $X=\{x_{1},x_{2},\ldots,x_{n}\}$ from an unknown density function $p(x)$, find an estimator for $p(x)$. We use the framework of maximum entropy to estimate $p(x)$ \cite{amari1982differential,csiszar2004information}. In the maximum entropy framework one is interested in identifying a unique distribution given a set of constraints on generalized moments of the distributions: $E_p[\phi(x)] = \alpha$ where $\phi(x)=[\phi_1(x),\phi_2(x),\ldots,\phi_m(x)]^T$ is an $m$-dimensional vector of basis functions defined over instance space ${\cal X}$, i.e., $\phi_i: {\cal X} \to \mathds{R}$ and $\alpha\in\mathds{R}^m$. {Note that the basis function $\phi_i(x)$, $i=\{1,2,\ldots,m\}$, can be any real valued function such as polynomials, splines or trigonometric series. For example, in a Gaussian distribution, $\phi_1(x)=x$ and $\phi_2(x) = x^2$. Additionally, $\alpha_i$ is the expected value of the $i$th feature function. For example, in a Gaussian distribution, $\alpha_1=\mu$ and $\alpha_2 = \sigma^2+\mu^2$.} This framework has the advantage of not restricting the class of the distribution to a specific density and considers a wide range of density functions in the class of exponential family hence, has a good approximation capability. In fact, it is shown that with a rich set of basis function $\phi(x)$, the approximation error decreases in order of $\mathcal{O}(1/m)$ where $m$ is the number of basis \cite{behmardi2011entropy}. We explain the maximum entropy approach below.

\subsection{Single density estimation (SDE)}
ME framework for density estimation was first proposed by Janes \cite{jaynes1957information} and has been applied in many areas of computer science and signal processing including natural language processing \cite{berger1996maximum}, species distribution modeling \cite{dudik2007maximum}, text classification \cite{zhu2005multi}, and image processing\cite{skilling1984maximum}. ME framework finds a unique probability density function $p(x)$ over $\mathcal{X}$ that satisfies the constraints $E_p[\phi(x)] = \alpha$. In principle, many density functions can satisfy the constraints. The maximum entropy approach selects a unique distribution among them that has the maximum entropy. The problem of single density estimation in the maximum entropy framework can be formulated as:

\begin{eqnarray}\label{eq:CCMaxEnt1a1}
&&\textrm{maximize} ~~~H(p)\\\nonumber
&&\textrm{subject to} ~~~ E_p[\phi]=\alpha\\\nonumber
&&\qquad\qquad\quad\int_{\mathcal{X}} p(x)dx=1,
\end{eqnarray}
where $H(p) = -\int p(x)\log p(x)dx$ is the entropy of $p(x)$ and $E_p[\phi(x)]=\int p(x)\phi(x) dx$. {For now, we ignore the non-negativity constraints and later show that the optimal solution, despite the exclusion of these constraints, is non-negative. Note that the objective function in (\ref{eq:CCMaxEnt1a1}) is strictly convcave (Hessian $-\frac{1}{p(x)}<0$) therefore, it has a unique global optimum solution. To find the optimum solution to (\ref{eq:CCMaxEnt1a1}) $p^*(x)$, we first construct the Lagrangian $\mathcal{L}(p(x),\lambda,\gamma)=H\left(p(x)\right) + \lambda^T ( E_p[\phi] - \alpha)-\gamma(\int p(x)dx-1)$, where $\lambda\in\mathds{R}^m$ and $\gamma\in\mathds{R}$ are dual variables corresponding to $m+1$ constraints imposed on the solution. Then, we hold $\lambda$ and $\gamma$ constant and maximize the Lagrangian w.r.t. $p(x)$. This yields an expression for $p^*(x)$ in terms of the dual variables. Finally, we substitute $p^*(x)$ back into the Lagrangian, solving for the optimal value of $\lambda$ and $\gamma$ ($\lambda^*$ and $\gamma^*$, respectively).}

{The derivative of the Lagrangian w.r.t. to $p(x)$ given $\lambda$ and $\gamma$ are fixed is given by
\begin{eqnarray}
\frac{\partial \mathcal{L}}{\partial p(x)} = -(\log p(x) + 1) + \lambda^T\phi(x)-\gamma.
\label{eq:derivationLagrangianPx}
\end{eqnarray}
}
{Equating (\ref{eq:derivationLagrangianPx}) to zero and solving for $p(x)$, we obtain
\begin{equation}
p^*(x;\lambda,\gamma) = \exp{\left(\lambda^T\phi(x)\right)}\exp{\left(-\gamma - 1\right)}.
\label{eq:pStarX}
\end{equation}
Equation (\ref{eq:pStarX}) represents $p^*(x)$ w.r.t. the dual variables $\lambda$ and $\gamma$. We can integrate $p^*(x)$ and set it to one to obtain $\gamma^*$. Hence, $\gamma^* = Z(\lambda)-1$, where $Z(\lambda) = \log\int_{\mathcal{X}} \exp{\{\lambda^T\phi(x)\}}dx$. Therefore, $p^*(x)$ can be written as
\begin{eqnarray}
p_\lambda^*(x) = \exp{\left(\lambda^T\phi(x)-Z(\lambda)\right)}.
\label{eq:pStarXFinal}
\end{eqnarray}
Note that the maximum entropy distribution given in (4), satisfies the non-negativity, since the exponential function is always non-negative. If we substitute $p_\lambda^*(x)$ and $\gamma^*$ back into the Lagrangian, we obtain the dual function $g(\lambda)$ as
\begin{eqnarray}
g(\lambda)&=&\mathcal{L}(p_\lambda^*(x),\lambda,\gamma^*)\nonumber\\
&=&Z(\lambda)-\lambda^T\alpha
\label{eq:dualME}
\end{eqnarray}
Therefore,
\begin{eqnarray}
\lambda^* = \arg\min_{\lambda} g(\lambda)
\label{eq:lambdaStar}
\end{eqnarray}
Since $g(\lambda)$ is a smooth convex function, setting $g(\lambda)=0$ yeilds optimum solution $\lambda^*$. Note that for $\lambda^*, \alpha=\frac{\partial Z(\lambda^*)}{\partial \lambda^*}$. Based on the fundamental principle in the theory of Lagrangian multiplier called, Karush-Kuhn-Tucker theorem, which asserts under the conditions that the cost function is convex and all the equality constraints are affine (both holds for (\ref{eq:CCMaxEnt1a1})), the optimal solution for dual is equal to the primal optimal solution \cite{boyd2004convex}. In other words, $p_\lambda^*(x)$ of (\ref{eq:pStarX}) with $\lambda=\lambda^*$ given by (\ref{eq:lambdaStar}) is in fact the optimum solution to (\ref{eq:CCMaxEnt1a1}). There is a one-to-one mapping between $\alpha$ in (\ref{eq:CCMaxEnt1a1}) and $\lambda$ in (\ref{eq:lambdaStar}). Therefore, $\{p(x|\alpha)|\alpha \in \mathds{R}^m\} = \{p(x|\lambda)|\lambda \in \mathds{R}^m\}$.
For example, consider $\mathcal{X}=\mathds{R}$ and $\phi(x)=[x\quad x^2]^T$. By (\ref{eq:pStarXFinal}), $p(x)=\exp{\left(\lambda_1x+\lambda_2x^2-Z(\lambda)\right)}$, where $Z(\lambda)= -\frac{\lambda_1^2}{4\lambda_2}+\log\sqrt{\frac{\pi}{-\lambda_2}}$, $\lambda_2\le 0$ or $p(x)=\frac{1}{\sqrt{2\pi\frac{1}{-2\lambda_2}}}\exp{\left(-\frac{(x-(-\frac{\lambda_1}{2\lambda_2}))^2}{2(\frac{1}{-2\lambda_2})}\right)}$. Given $p(x;\lambda_1,\lambda_2)$, $E[\phi(x)]=[-\frac{\lambda_1}{2\lambda_2}\quad \frac{1}{-2\lambda_2}+\frac{\lambda_1^2}{4\lambda_2^2}]^T=[\alpha_1\quad \alpha_2]^T$.}

We will now derive the maximum-likelihood (ML) estimator for the parameter $\lambda$ in $p_{\lambda}(x)$ given $n$ $i.i.d.$ observations $x_1, \ldots, x_n$. The log likelihood for (\ref{eq:pStarXFinal}) can be written as
\begin{eqnarray}\label{eq:CCMaxEntlik}
&&\mathcal{L}(\lambda) = \log p(x_1,x_2,\ldots,x_n) = \sum_{i=1}^n (\lambda^T\phi(x_i)-Z(\lambda)) \nonumber\\
&&~~~~~~= n (\lambda^TE_{\hat{p}}[\phi(x)]-Z(\lambda)),
\end{eqnarray}
where $E_{\hat{p}}[ g(x)]$ denotes the empirical average of $g(x)$ given by $\frac{1}{n} \sum_{i=1}^n g(x_i)$. Thus, we can write the negative log-likelihood function as follows:
\begin{eqnarray}\label{eq:CCMaxEntkl}
-\mathcal{L}(\lambda) &=& -n E_{\hat{p}}[\lambda^T\phi(x)-Z(\lambda)]\nonumber\\
& = & nE_{\hat{p}}[\log \hat{p}] - n E_{\hat{p}}[\lambda^T\phi(x)-Z(\lambda)]-nE_{\hat{p}}[\log \hat{p}]\nonumber\\
&=& nD(\hat{p}\|p_{\lambda}) + \Upsilon,
\end{eqnarray}
where $\Upsilon = -nE_{\hat{p}}[\log \hat{p}]$ is a constant w.r.t. $\lambda$ and $D(p\|q) = E_p[\log p - \log q]$. Therefore, maximizing the log-likelihood in (\ref{eq:CCMaxEntlik}) w.r.t. $\lambda$ is equivalent to minimizing the KL-divergence in (\ref{eq:CCMaxEntkl}) w.r.t. $\lambda$. Thus, $\hat\lambda$ can be obtained as a result of the following optimization problem:
\begin{eqnarray}\label{eq:MLEstimator}
\hat\lambda & = & \arg\min_{\lambda} nD(\hat{p}\|p_{\lambda})\nonumber\\
& = & \arg\min_{\lambda} n(Z(\lambda)-\lambda^T E_{\hat{p}}[\phi(x)]).
\end{eqnarray}
There are several algorithms for solving ME, e.g., iterative scaling  \cite{della1997inducing} and its variants \cite{dudik2007maximum}, gradient descent, Newton, and quasi-Newton approach \cite{salakhutdinov2003convergence}. The ML optimization problem is convex in terms of $\lambda$ and can be solved efficiently using Newton's method. Newton's method requires the first and second derivative of the objective function w.r.t. $\lambda$. The derivatives of $nD(\hat{p}\|p_{\lambda})$ are:
\begin{eqnarray*}\label{eq:CCMaxEntdefs}
&\nabla_\lambda =  n(E_{p_\lambda}[\phi]-E_{\hat{p}}[{\phi}])\nonumber\\
&\nabla^2_\lambda =  n(E_{p_\lambda}[\phi]E_{p_\lambda}[\phi]^T-E_{p_\lambda}[\phi\phi^T]).
\end{eqnarray*}
Algorithm~\ref{al:CCMaxEntalgorithmI} provides the details for Newton's method implementation.
\begin{algorithm}[!ht]
\caption{Single density estimation algorithm}
\begin{algorithmic}
\STATE Input: $X=\{x_{1},x_{2},\ldots,x_{n}\}$ samples from bag $X$, $K$, $\phi\in \mathds{R}^m$, $\lambda^0\in\mathds{R}^m$.
\STATE Output: $\hat\lambda\in\mathds{R}^m$
\FOR{$k = 1$ to $K$}
\STATE $\Delta \lambda^k = - {\nabla^2 _{\lambda^{k}}}^{-1}\nabla_\lambda^k$
\STATE Find $t^k$ using backtracking
\STATE $\lambda^{k+1} = \lambda^{k} + t^k\Delta\lambda^k$
\ENDFOR
\STATE $\hat\lambda = \lambda^K$
\end{algorithmic}
\label{al:CCMaxEntalgorithmI}
\end{algorithm}
ME can overfit data due to low number of samples or large number of basis functions $\phi(x)$\cite{dudik2007maximum}. Regularized ME (RME) is proposed to overcome the issue of overfitting in ME \cite{krishnapuram2005sparse,dudik2004performance}. RME can be either formulated as relaxing the equality in (\ref{eq:CCMaxEnt1a1}) or putting a prior on the p.d.f. in (\ref{eq:CCMaxEnt1a1}) \cite{chen2000survey} (Laplace prior yields $l_1$ regularization and Gaussian prior yields $l_2$ regularization).  Algorithms for solving RME are proposed in \cite{chen2000survey,dudik2007maximum}. Convergence analysis for RME is provided in \cite{behmardi2011entropy,dudik2007maximum}. The problem of single density estimation is presented to introduce the maximum entropy framework for density estimation. In the next section, the principle of maximum entropy is applied to multiple density estimation.

\subsection{Multiple density estimation (MDE)}
Multiple density estimation (MDE) for multi-instance data can be done following the same principle as explained for single density estimation in the previous section. In MDE each bag is represented by one distribution, i.e., $p_i(x) = p_{\lambda_i}(x) =\exp{\left(\lambda_i^T\phi(x) - Z(\lambda_i)\right)}$ and the cost function for MDE, due to bag independence, is the sum of the individual bags negative log-likelihood. MDE can be solved using the following minimization:
\begin{eqnarray}\label{eq:CCMaxEntmd}
{\hat\Lambda} &=& \arg\min_{{\Lambda}} \sum_{i = 1}^N n_i D(\hat{p_i}\|p_{\lambda_i})\nonumber\\
& = & \arg\min_{{\Lambda}}\sum_{i=1}^{N}n_i(Z(\lambda_i)-\lambda_i^T E_{\hat{p}}[\phi_i]),
\end{eqnarray}
where $\hat\Lambda = [\hat\lambda_1,\ldots,\hat\lambda_N]$, $\Lambda = [\lambda_1,\ldots,\lambda_N]$, $N$ is total number of bags, and $n_i$ is total number of instances in the $i$th bag. The objective function in (\ref{eq:CCMaxEntmd}) is expressed as the sum of the functions of individual variables $\lambda_i$ which only depends on the parameters of single density. Hence, MDE formulation proposed in (\ref{eq:CCMaxEntmd}) considers the density estimation for each bag individually. This individual estimate addresses the nature of each dataset separately and ignores the fact that the underlying structure of the data can be shared among all datasets. This might cause a poor generalization performance due to the low number of samples for some bags \cite{dudik2007hierarchical}. To address this, we use a joint regularization on the parameter space to simultaneously learn the
structure of the distribution space and infer each distribution while keep the origin of each data uninfluenced. Hierarchical density estimation \cite{dudik2004performance} formulates the problem of MDE using $l_1$ regularization. The regularization defined on each data separately and on the group of the data defined in the hierarchy. Note that the hierarchical structure of the data is a prior information. However, in most cases in real world applications the relations among the datasets are unknown beforehand, e.g., in text or image datasets. In the following, we proposed a framework for learning jointly in the space of distributions using the principle of maximum entropy.
\section{Structured multiple density estimation}
To improve the power of estimation in maximum entropy framework, we choose a large number of basis functions $\phi$. This results in a large dimensional space for the parameter $\lambda$ which cause overfitting and poor generalization performance. To reduce the effect of overfitting for density estimation, an efficient way is to reduce the dimensionality of the parameter space. This low dimensional space corresponds to the hidden structure of the data. Rank minimization is an approach in dimension reduction which finds a linear subspace of the observed data by constraining the dimension of the given matrix. Rank minimization introduces structures in the parameter space. In the following we first define rank recovery in the space of distributions and then show how it can be formulated in the maximum entropy framework.
\subsection{Rank recovery in the space of distributions}
The dimension of the space of distributions is controlled by the size of the basis $\phi = [\phi_1,\phi_2,\ldots,\phi_m]^T$. Often the size of $\phi$ is large to allow accurate approximation of the distribution space. Hence, we are interested in finding a smaller basis that provides a fairly accurate replacement to the original basis $\phi$. We consider the problem of finding a new basis in the span of $\phi$. Suppose a smaller basis $\psi$ can be obtained by $\psi = A^T\phi$, where $\psi = [\psi_1,\psi_2,\ldots,\psi_k]^T$ and A is a $m \times k$ matrix, where $k < m$. Instead of using $\lambda_i^T\Phi$ involving $m$ terms, one can use $\beta_i^T\psi$ involving only $k$ terms. In this case, $\phi^T\Lambda = \psi^T\beta=\phi^TA\beta$, where $\Lambda = [\lambda_1,\lambda_2,\ldots,\lambda_N]$ and $\lambda_i \in \mathds{R}^m$, which results in $\Lambda = A\beta$ such that $A \in\mathds{R}^{m\times k}$ and $\beta \in \mathds{R}^{k\times N}$. Hence $\Lambda = A\beta$ is a low-rank matrix.

\subsection{Regularized MDE (RMDE) using maximum entropy}
To obtain a low-rank solution for $\Lambda$, we can solve a regularized nuclear norm MDE. The nuclear norm of a matrix $\|X\|_*$ is defined as the sum of the singular values of matrix $X$. The nuclear norm is a special class of Schatten norm which is defined as $\|X\|_p = (\sum_{i} \sigma_i^p)^\frac{1}{p}$. When $p=1, \|X\|_p$ is equal to the nuclear norm. Nuclear norm enforces sparsity on the singular values of matrix $X$, which results in a low-rank structure. The heuristic replacement of rank with nuclear norm has been proposed for various applications such as matrix completion \cite{recht706guaranteed,candes2010matrix}, collaborative filtering\cite{srebro2005maximum}, and multi-task learning\cite{pong2009trace}.

In RMDE, a regularized nuclear norm is added to the objective function in (\ref{eq:CCMaxEntmd}) yielding:
\begin{eqnarray}\label{eq:CCMaxEnt20}
&&\textrm{minimize}\qquad \sum_{i = 1}^N n_i(Z(\lambda_i)-\lambda_i^T E_{\hat{p}}[\phi_i]) + \eta \|\Lambda\|_*,
\end{eqnarray}
where $\eta$ is the regularization parameter. RMDE can be viewed as maximum a posteriori (MAP) criterion using a prior distribution over matrix $\Lambda$ of the form $Ce^{-\eta\|\Lambda\|_*}$. This is similar to the interpretation of $l_1$-regularization for sparse recovery as MAP with a Laplacian prior. Recently, We proposed a quasi-Newton approach to solve RMDE \cite{behmardissp2}. RMDE can also be formulated as a constrained MDE as follows:
\begin{eqnarray}\label{eq:CCMaxEnt21}
&&\textrm{minimize}\qquad \sum_{i = 1}^N n_i(Z(\lambda_i)-\lambda_i^T E_{\hat{p}}[\phi_i]),\nonumber\\
&& \textrm{subject to} \qquad\|\Lambda\|_* \le \nu,
\end{eqnarray}
where $\nu\ge 0$ is a tuning parameter. For each value of $\eta$ in (\ref{eq:CCMaxEnt20}) there is a value of $\nu$ in (\ref{eq:CCMaxEnt21}) which produces the same solution \cite{gill1981practical}. One of the main challenges in regularized and constrained MDE is the choice of regularization parameters $\eta$ and $\nu$. Often, the regularization parameter is chosen based on cross-validation which is computationally demanding and is always biased toward the noise in the validation set. There is an extensive discussion in \cite{behmardiTSP} for model selection in topic model. We propose the concept of confidence-constrained rank minimization for jointly learning the space of distributions which overcome the issues of parameter tuning with regularized and constrained MDE.
\section{Confidence-constrained maximum entropy (CME)}
We propose the framework of confidence-constrained maximum entropy (CME) for learning from multi-instance data. The difficulties in tuning the regularization parameters in regularized MDE will be addressed in CME by solving a constrained optimization problem where the constraint only depends to the dimension of the data (i.e., number of instances and number of bags). Using the properties of the maximum entropy framework, an in-probability bound on the objective function in (\ref{eq:CCMaxEntmd}) can be obtained. The probability bound on the log-likelihood function allows us to define a confidence set. A confidence set is a high-dimensional generalization of the confidence interval that we use to restrict the search space of the problem. Search for a low-rank $\Lambda$ inside the confidence set guarantees a low-rank solution with high probability. Hence, in this approach the roles of ML objective and rank constraint are reversed. We consider rank minimization subject to ML objective constraint. The CME is given by:
\begin{eqnarray}\label{eq:CCMaxEntrank}
&&\textrm{minimize}\qquad \textrm{Rank}(\Lambda)\nonumber\\
&& \textrm{subject to} \qquad\sum_{i = 1}^N n_iD(p_{\hat{\lambda}_i}\|p_{\lambda_i}) \le \epsilon(\omega_a),
\end{eqnarray}
where $\epsilon(\omega_a)=\frac{aNm}{2}$ is an in-probability bound for the estimation error. $N$ is total number of bags and $m$ is total number of feature functions $\phi$. Note in this formulation the tuning parameter $\epsilon(\omega_a)$ can be obtained by bounding $\sum_{i = 1}^N n_i D(p_{\hat\lambda_i}\|p_{\lambda_i})$  using the following theorem.
\begin{theorem}\label{th:th1}
Let $\hat{\lambda}=arg\min_{\lambda} nD(\hat{p}\|p_{\lambda})$ defined in (\ref{eq:MLEstimator}). With probability at least $1-\omega_a$:
\begin{eqnarray*}
p\biggl( \sum_{i=1}^Nn_iD(p_{\hat{\lambda}_i}\|p_{\lambda_i})\ge \epsilon(\omega_a)\biggl) \le \frac{1}{a}.
\end{eqnarray*}
\end{theorem}
(For proof, see Appendix~\ref{appen:1}). This theorem suggests that the original low-rank representation distributions associated with the $N$ bags $p_{\lambda_1}$, $p_{\lambda_2}$, and $p_{\lambda_N}$ can be found within an $\epsilon(\omega_a)$-ball (as in (\ref{eq:CCMaxEntrank})) around the rank-unrestricted ML estimates $p_{\hat{\lambda}_1}$, $p_{\hat{\lambda}_2}$, and $p_{\hat{\lambda}_N}$ with high probability. Additionally, $\epsilon(\omega_a)$ is free of any tuning parameters. It only depends on the dimensions of dataset which is available prior to observing the data.  Since (\ref{eq:CCMaxEntrank}) involves rank minimization which is non-convex, we provide an alternative convex relaxation to (\ref{eq:CCMaxEntrank}) in the following.
\subsection{Confidence-constrained maximum entropy nuclear norm minimization  (CMEN)}
Constrained rank recovery of an unknown matrix has been studied extensively in the literature in the communities of signal processing, control system, and machine learning in problems such as matrix completion and matrix decomposition \cite{candes2009robust}. In general, rank minimization problems are NP hard \cite{meka2008rank}. Various algorithms have been proposed to solve the general rank minimization problem locally (e.g., see \cite{haldar2009rank}). To solve the rank minimization problem proposed in (\ref{eq:CCMaxEntrank}), we propose to apply the widely adopted approach of replacing the rank minimization with the tractable convex optimization problem of nuclear norm minimization. In the following, CME nuclear norm minimization is proposed as a convex alternative to (\ref{eq:CCMaxEntrank}):
\begin{eqnarray}\label{eq:CCMaxEnt24}
&&\textrm{minimize}~~~~ \|\Lambda\|_*\nonumber\\
&& \textrm{subject to}~~~~\sum_{i = 1}^N n_i D(p_{\hat{\lambda}_i}\|p_{\lambda_i}) \le \epsilon.
\end{eqnarray}
We denote the solution to (\ref{eq:CCMaxEnt24}) by $\hat\Lambda_*$. Since the nuclear norm is a convex function, and the set of the inequality and equality constraints construct a convex set, (\ref{eq:CCMaxEnt24}) is a convex optimization problem. This nuclear norm regularization encourages a low-rank representation to feature space, i.e., all features can be represented as a linear combination of a few alternative features. Consider the singular value decomposition of $\Lambda = USV^T = \sum_j u_j s_j v_j^T$, then
\begin{eqnarray*}\label{eq:CCMaxEntpsi}
\lambda_i^T \phi(x) &=& \sum_{j=1}^k s_j (e_i^T v_j) (u_j^T \phi(x))\nonumber\\
& = & \sum_{j=1}^k s_j (e_i^T v_j) \psi_j(x) = \beta_i^T\psi(x)
\end{eqnarray*}
where $k$ is the rank of matrix $\Lambda$. Similar to principle component analysis, where each data point can be approximated as a linear combination of a few principle components, each bag can be represented as a distribution using a linear combination of only a few basis functions $[\psi_1(x),\ldots,\psi_k(x)]$. This method facilitates a dimension reduction in the space of distributions by representing each distribution with a lower number of basis functions $\psi$ ($k \ll m$).
\section{Confidence-constrained maximum entropy nuclear norm minimization algorithm (CMENA)}
The optimization problem in (\ref{eq:CCMaxEnt24}) can be written as follows:
\begin{eqnarray}\label{eq:CCMaxEnt25}
&&\textrm{minimize}~~~~ f(\Lambda)\nonumber\\
&& \textrm{subject to}~~~~g(\Lambda) \le \epsilon,
\end{eqnarray}
where $f(\Lambda)=\|\Lambda\|_*$ and $g(\Lambda) = \sum_{i = 1}^N n_iD(p_{\hat{\lambda}_i}\|p_{\lambda_i})$. The Lagrangian of (\ref{eq:CCMaxEnt25}) is
\begin{eqnarray}\label{eq:CCMaxEnt26}
\mathcal{L}(\Lambda,z) = f(\Lambda) + z(g(\Lambda)-\epsilon),
\end{eqnarray}
where $z\ge 0$ is the Lagrangian multiplier. The next step is to minimize the Lagrangian (\ref{eq:CCMaxEnt26}) with respect to the primal variable $\Lambda$. Define $\Lambda^*(z)$ as:
\begin{eqnarray}\label{eq:Lagrangian_dual_optimization_wrt_primal}
\Lambda^*(z) = \arg\min_{\Lambda} \mathcal{L}(\Lambda,z).
\end{eqnarray}
By replacing $\Lambda^*(z)$ in the Lagrangian (\ref{eq:CCMaxEnt26}), we obtain the dual:
\begin{eqnarray*}
y(z) = \mathcal{L}(\Lambda^*(z),z).
\end{eqnarray*}
The dual formulation is given by the following optimization
\begin{eqnarray*}
\textrm{maximize}~~~ y(z)\nonumber\\
\textrm{subject to} ~~ z\ge 0.
\end{eqnarray*}
To optimize the Lagrangian with respect to the primal variable $\Lambda$, we propose to use the proximal gradient approach. In the following, we introduce the proximal gradient algorithm and then show how it can be applied to solve (\ref{eq:Lagrangian_dual_optimization_wrt_primal}).
\subsection{Proximal gradient algorithm}
Consider a general unconstrained nonsmooth convex optimization problem in the form of the following:
\begin{eqnarray}\label{eq:general_equation_pga}
\textrm{minimize}~~~ P(X)\coloneqq f(X) + g(X),
\end{eqnarray}
where $f:\mathds{R}^{m\times n} \to \mathds{R}$ is a convex, lower semicontinuous (lsc) \cite{liu2009implementable} function and $g:\mathds{R}^{m\times n}\to \mathds{R}$ is a smooth convex function (i.e., continuously differentiable). Assume $\nabla g(X)$ is Lipschitz continuous on the domain of $g$, i.e.,
\begin{eqnarray*}
\|\nabla g(X) - \nabla g(Y) \|_F \le \tau_g \|X-Y\|_F,~~\forall X,Y \in \mathds{R}^{m\times n},
\end{eqnarray*}
where $\tau_g >0$ is some positive scalar. Therefore, a quadratic approximation of $g$ at point $X_0$ can be provided as follows:
\begin{eqnarray*}
g(X) &\le& g(X_0) +  \langle X-X_0,\nabla g(X_0)\rangle + \frac{\tau_g}{2}\|X-X_0\|_F^2.
\end{eqnarray*}
Instead of minimizing $P(X)$ in (\ref{eq:general_equation_pga}), we minimize an upper bound on $P(X)$, i.e.,
\begin{eqnarray*}
&&P(X) \le f(X) + g(X_0) + \langle X-X_0,\nabla g(X_0)\rangle\nonumber\\
&& + \frac{\tau_g}{2}\|X-X_0\|_F^2\nonumber\\
&&= Q(X,X_0),
\end{eqnarray*}
where $Q(X,X_0)$ is $f(X)$ plus a simple quadratic local model of $g(X)$ around $X_0$.

To proceed further, we need to define the proximal mapping (operator). A proximal mapping is an operator defined for a convex function $h$ as $\textrm{prox}_h(x) = \arg\min_u~ h(u) + \frac{1}{2}\|x-u\|_2^2$. For example, if $h(x)$ is the indicator function of set $C$ the proximal mapping is the projection into set $C$ and if $h(x)$ is $\|\cdot\|_1$ the proximal mapping is the soft thresholding operator \cite{liu2009implementable}.

Since $Q(X,X_0)$ can be reformulated as
\begin{eqnarray*}
&&Q(X,X_0) = f(X) + \frac{\tau_g}{2}\|X - (X_0 - \frac{1}{\tau_g}\nabla g(X_0))\|_F^2 +g(X_0)\nonumber\\
&&~~~~-\frac{1}{2\tau_g}\|\nabla g(X_0)\|_F^2,
\end{eqnarray*}
then $X^*$ the minimum of $Q$ is
\begin{eqnarray*}
X^* &=& \arg\min Q(X,X_0)\nonumber\\
&=& \arg\min f(X) + \frac{\tau_g}{2}\|X - (X_0 - \frac{1}{\tau_g}\nabla g(X_0))\|_F^2\nonumber\\
& = & \Pi(X').
\end{eqnarray*}
where $X' = X_0 - \frac{1}{\tau_g}\nabla g(X_0)$. The proximal operator $\Pi(X)$ is given by $\Pi(X) = \arg\min_Y f(Y) +\frac{1}{2}\|X-Y\|_F^2$. Moreover, it can be found in closed form for some nonsmooth convex functions (e.g., nuclear norm) which is an advantage of algorithm to solve large scale optimization problem \cite{toh2010accelerated}. Note that if $f(X)=0$ then $X^{\textrm{new}} = X^{\textrm{old}}+\frac{1}{\tau}\nabla g(X^{\textrm{old}})$, i.e., the proximal gradient algorithm reduces to the standard gradient algorithm. The convergence rate for the proximal gradient algorithm is $\mathcal{O}(1/k)$ where $k$ is the number of iterations (i.e., see \cite{toh2010accelerated}~Theorem~2.1).
\subsection{Proximal gradient algorithm to solve CMEN}
Given $\nabla g(\Lambda)$ is Lipschitz continuous with parameter $\tau_g=Nm$ (see Appendix~{\ref{app:lipschitz1}), where $N$ is total number of bags and $m$ is total number of feature functions, a quadratic upper bound for (\ref{eq:CCMaxEnt26}) can be written as:
\begin{eqnarray*}
&&\mathcal{L}(\Lambda,z) \le \|\Lambda\|_* +  z(\frac{\tau_g}{2}\|\Lambda-\Lambda'\|_F^2+g(\Lambda_0)\nonumber\\
&&-\frac{1}{2\tau_g}\|\nabla g(\Lambda_0)\|_F^2-\epsilon)\nonumber\\
&&= Q(\Lambda,\Lambda_0)
\end{eqnarray*}
where $\Lambda' = \Lambda_0-\frac{1}{\tau_g}\nabla g(\Lambda_0)$. The solution to the minimization of $Q(\Lambda,\Lambda_0)$ w.r.t. $\Lambda$ is
\begin{eqnarray*}
\hat\Lambda_*(z) &=& \arg\min Q(\Lambda,\Lambda_0)\nonumber\\
&=& \mathcal{D}_{\frac{1}{\tau_gz}}(\Lambda')
\end{eqnarray*}
where $\mathcal{D}_\alpha(X)$ is the soft-thresholding operator on the singular values of matrix $X$ (for proof see \cite{cai810singular}) defined by $D_\alpha(X) = U(S-\alpha I)_+V^T$, where $X=USV^T$ is the SVD of $X$. To find $z^*$ we have to maximize $Q(\hat\Lambda_*(z),\Lambda_0)$ w.r.t. $z$. Since parameter $z$ is a scalar, we propose a greedy search approach to find the optimum $z$ (see Algorithm~\ref{al:CCMaxEntalgorithmIII}).
\subsection{step size}
In the proximal gradient approach, $\Lambda$ will be updated in each iteration based on $1/\tau_g$. In fact, $1/\tau_g$ plays the role of step size. However, in practice it is usually very conservative to set a constant step size $\tau_g$ \cite{toh2010accelerated}. As long as the inequality $L(\Lambda,z) \le Q(\Lambda,\Lambda_0)$ is hold, the step size can be increased. Therefore, a linesearch-like algorithm is proposed to find a smaller value for $\tau_g$ which satisfies the inequality (see Algorithm~\ref{al:CCMaxEntalgorithmIII}).
The pseudo code for CMENA is proposed in Algorithm~\ref{al:CCMaxEntalgorithmIII}.
\begin{algorithm}[!ht]
\caption{CMENA}
\begin{algorithmic}
\STATE Input: $X_i=\{x_{i1},x_{i2},\ldots,x_{in_i}\}$ sample from bag $X_i$, $i=1,\ldots,N$, $K$, $\phi\in \mathds{R}^m$, $\Lambda^1,\Lambda^0\in\mathds{R}^{m\times N}$,$a_1=a_0=1$, $z^1_{-}$, $z^1_{+}$, and $\alpha\in(0,1)$.
\STATE Output: $\lambda_i^*\in\mathds{R}^{m}$ and $Z(\lambda_i^*)$
\FOR{$j = 1$ to $\ldots$}
\STATE $z^k = \frac{z^k_{-}+z^k_{+}}{2}$ \COMMENT{Dual variable update}
\FOR{$k = 1$ to $K$}
\STATE $\bar\Lambda^k = \Lambda^k + \frac{a^{k-1}-1}{a^{k}}(\Lambda^{k} - \Lambda^{k-1})$\COMMENT{Acceleration}
\WHILE{$L(\bar\Lambda^k,z^k) \le Q(\bar\Lambda^k,\bar\Lambda^{k-1}$)}
\STATE $\tau_g^k = \alpha\tau_g^{k-1}$ \COMMENT{Line search}
\ENDWHILE
\STATE $G^k =  \Lambda^k - \frac{1}{\tau_g^k}\nabla g(\bar\Lambda^k)$ \COMMENT{Variable update}
\STATE Compute $\Lambda^{k+1} = D_\frac{1}{z^k\tau_g^k}(G^k)$ \COMMENT{proximal operator}
\STATE $a^{k+1} = \frac{1+\sqrt{1+4{a^{k}}^2}}{2}$
\ENDFOR

\COMMENT{Line search for dual variable $z$}
\IF{$g(\Lambda)-\epsilon \ge 0$ }
\STATE $z^{k+1}_{-} = z_k$
\ELSE
\STATE $z^{k+1}_{+} = z_k$
\ENDIF
\IF{$\sum_iD(p_{\hat{\lambda}_i}\|p_{\lambda_i})n_i - \epsilon<\emph{consTol}$}
\STATE break
\ENDIF
\ENDFOR
\end{algorithmic}
\label{al:CCMaxEntalgorithmIII}
\end{algorithm}

\subsection{Acceleration}
The convergence rate for the proximal gradient approach is $\mathcal{O}(1/k)$ where $k$ is the number of iteration \cite{toh2010accelerated,liu2009implementable}. The convergence rate of the gradient approach can be speed up to $\mathcal{O}(1/k^2)$ using the extrapolation technique proposed in \cite{nesterov1983method} given the fact that the $\nabla g(\Lambda)$ is Lipschitz continuous with $\tau_g = Nm$ (see Appendix~\ref{app:lipschitz1}). We define the extrapolated solution as follows:
\begin{eqnarray*}
\bar\Lambda^k = \Lambda^k + \frac{a^{k-1}-1}{a^{k}}(\Lambda^{k} - \Lambda^{k-1}),
\end{eqnarray*}
where $a_k = \frac{1+ \sqrt{1+4a^{2k}}}{2}$.
The only costly part of the proximal algorithm is the evaluation of the singular values in each iteration. Note that in each iteration of soft-thresholding operator we need to know the number of singular values greater than a threshold. As in \cite{toh2010accelerated,cai810singular,lin2009augmented,behmardiTSP}, we use the PROPACK package to compute a partial SVD. Because PROPACK can not automatically calculate the singular values which are greater than specific value $\zeta$, we use the following procedure. To facilitate the computation of singular value $5$ at a time, we set $b_0 = 5$ and update $b_{l+1}$ for $l = 0,1,\ldots$ as follows:
\begin{eqnarray*}
b_{l+1} = \left\{\begin{array}{cc}
\textrm{Rank}(\Lambda^{k+1}) & \textrm{if Rank}(\Lambda^{k+1}) < b_k\\
\textrm{Rank}(\Lambda^{k+1}) + 5 & \textrm{if Rank}(\Lambda^{k+1}) \ge b_k.\\
\end{array}\right.
\end{eqnarray*}
This procedure stops when $b_{l+1} = b_l$. Partial SVD calculation reduces the cost of the computation significantly, especially in the low-rank setting. The pseudo code for calculating SVD is in Algorithm~\ref{al:algorithmIII}.
\begin{algorithm}[!ht]
\caption{SVD calculation using PROPACK}
\begin{algorithmic}
\STATE Choose $r_0 = 0$, and $i=5$
\STATE in step $l$
\STATE $b_l = r_{k-1}+1$
\REPEAT
\STATE $[U S V]_{b_l} = \textrm{SVD}(\Lambda^k)$
\STATE $b_l = b_l + i$
\UNTIL{$s^k_{b_l-i} \le \frac{1}{z^k\tau_g^k}$}
\STATE $r_k = \max\{j: s_j^k > \frac{1}{z^k\tau_g^k}\}$
\STATE $\Lambda^{k+1}=\sum_{j=1}^{r_k}(s_j^k-\frac{1}{z^k\tau_g^k})u^k_jv^k_j$
\end{algorithmic}
\label{al:algorithmIII}
\end{algorithm}

\section{Experiments}\label{se:4}
In this section, we evaluate both theoretical and computational aspect of CMEN ans compare to RMDE for rank recovery in the space of distributions. For the theoretical part, we provide a phase diagram analysis to evaluate the performance of both CMEN and RMDE in exact rank recovery. We then provide an illustration of distribution space dimension reduction using CMEN. Moreover, we show that CMEN introduces a metric which can be used in object similarity recognition in image processing.
\subsection{Phase diagram analysis}
We use the notion of phase diagram \cite{donoho2006sparse} to evaluate probability of exact rank recovery using CMEN and RMDE for a wide range of matrices $\Lambda$ of different dimensions (i.e., features size $\times$ number of bags) and different values for the rank of matrix $\Lambda$. We construct distributions using low-rank matrix $\Lambda$ and draw $i.i.d$ samples using  rejection sampling (data are generated in $2D$ space). {For the basis functions used in constructing the maximum entropy distribution space, we propose $\phi_{2k} = \cos (g_k^T x)$ and $\phi_{2k-1}=\sin (g_k^T x)$, where $g_k \sim {\cal N} (0, I)$ i.i.d for $k=1,2,\ldots,m/2$. In \cite{rahimi2007random}, a similar transformation is used to approximate Gaussian kernels.} Figure~\ref{fig:122} shows the contour plot of the first $4$ distributions used in our experiments. For the random samples drawn from the constructed distributions, we obtain $\hat\Lambda$ by maximum likelihood estimation  (\ref{eq:CCMaxEntmd}). Note that $\hat\Lambda$ is a noisy version of matrix $\Lambda$ and is full rank.
\begin{figure}[ht]
\centering
\includegraphics[scale = 0.2]{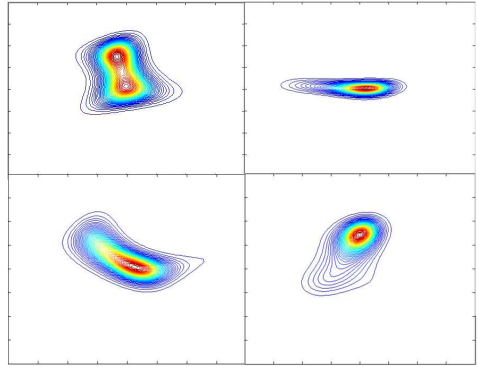}
\caption{Contour plot of the first $4$ distributions used in our experiment }\label{fig:122}
\end{figure}
We consider two different setups for number of bags: $N = 50$ and $N = 500$. We would like to illustrate the performance of CMEN and RMDE in small ($N = 50$) and large ($N = 500$) scale problems in terms of exact rank recovery. For $N=50$ bags, we vary the number of features and rank of matrix $\Lambda$ over a grid of $(m,T)$ with $m$ (number of features) ranging through $7$ equispaced points in the interval $[20,50]$ and $T$ (rank of matrix $\Lambda$) ranging through $10$ equispaced points in the interval $[2,20]$ (see Fig.~\ref{fig:55}). Each pixel intensity in the phase diagram corresponds to the empirical evaluation of the probability of exact rank recovery. For each pixel in the phase diagram we produce $10$ realization of $\hat\Lambda$.  We run CMEN and RMDE for each of $10$ realization of $\hat\Lambda$ and compare the rank of the obtained matrix $\Lambda^*$ with the rank of the true $\Lambda$. The rank evaluation is done by counting the number of singular values of matrix $\Lambda^*$ exceeding a threshold. The threshold is defined based on the empirical distribution of the smallest nonzero singular values of the true matrix $\Lambda$ (i.e., mean minus three times the standard deviation). To find the regularization parameter $\eta$ in RMDE (\ref{eq:CCMaxEnt20}), we consider both a cross-validation approach and a continuation technique \cite{toh2010accelerated,ma2011fixed}. The continuation technique in nuclear norm minimization is similar to the path following algorithm in solving $l_1$ regularized regression (LASSO) proposed in \cite{efron2004least}. Convergence analysis of the continuation technique is shown in \cite{hale2007fixed}. For cross-validation, we consider a range of regularization parameter $\eta = \{10^{-4},10^{-3},\ldots,10^3,10^4\}$. For each value of $\eta$, we separate data into training and test sets ($70\%$ training and $30\%$ test), and evaluate the test error using the objective function in (\ref{eq:CCMaxEnt20}), then select $\eta^*$ as the value corresponding to the lowest test error. For the continuation technique, we set $\eta$ to a large value $(\eta^0 = \|\hat\Lambda\|^2_F)$ and repeatedly solve the optimization problem (\ref{eq:CCMaxEnt20}) with a decreasing sequence of $\eta^k$ until we reach the target value $\bar\eta$ ($\eta^k = \max(10^{-1}\eta^{k-1},\bar\eta)$) where $\bar\eta = 10^{-3}\eta^0$. Due to large value of $\eta$ in the beginning of the algorithm, matrix $\Lambda^*$ is low-rank and in each iteration we increase the rank of $\Lambda^*$. Note that the value of constant $10^{-3}$ in $\bar\eta = 10^{-3}\eta^0$ and $10^{-1}$ in $\eta^k = \max(10^{-1}\eta^{k-1},\bar\eta)$ is set manually based on preliminary experiments. The stopping criterion for CMEN is the combination of \emph{MaxIter} $\le 100$, \emph{objTol} $< 10^{-2}$, and \emph{consTol} $< 10^{-1}$ where \emph{MaxIter} is the maximum number of iteration of main algorithm, \emph{objTol} is the tolerance of objective function $\|\textrm{f}^{k-1}_{\min} - \textrm{f}^{k}_{\min}\|_1$, and \emph{consTol} is the tolerance for violating the confidence constraint $\| \sum_iD(p_{\hat{\lambda}_i}\|p_{\lambda_i})n_i - \epsilon\|$. The stopping criteria for RMDE is the same as for CMEN except that \emph{consTol} is not used.  Figure~\ref{fig:55a}, \ref{fig:55b}, and \ref{fig:55c} show the phase diagram results for exact rank recovery with CMEN, RMDE (cross-validation), and RMDE (continuation technique) for $N = 50$.
\begin{figure}[ht]
\centering
\subfigure [] {
\resizebox{4cm}{!}{
\includegraphics{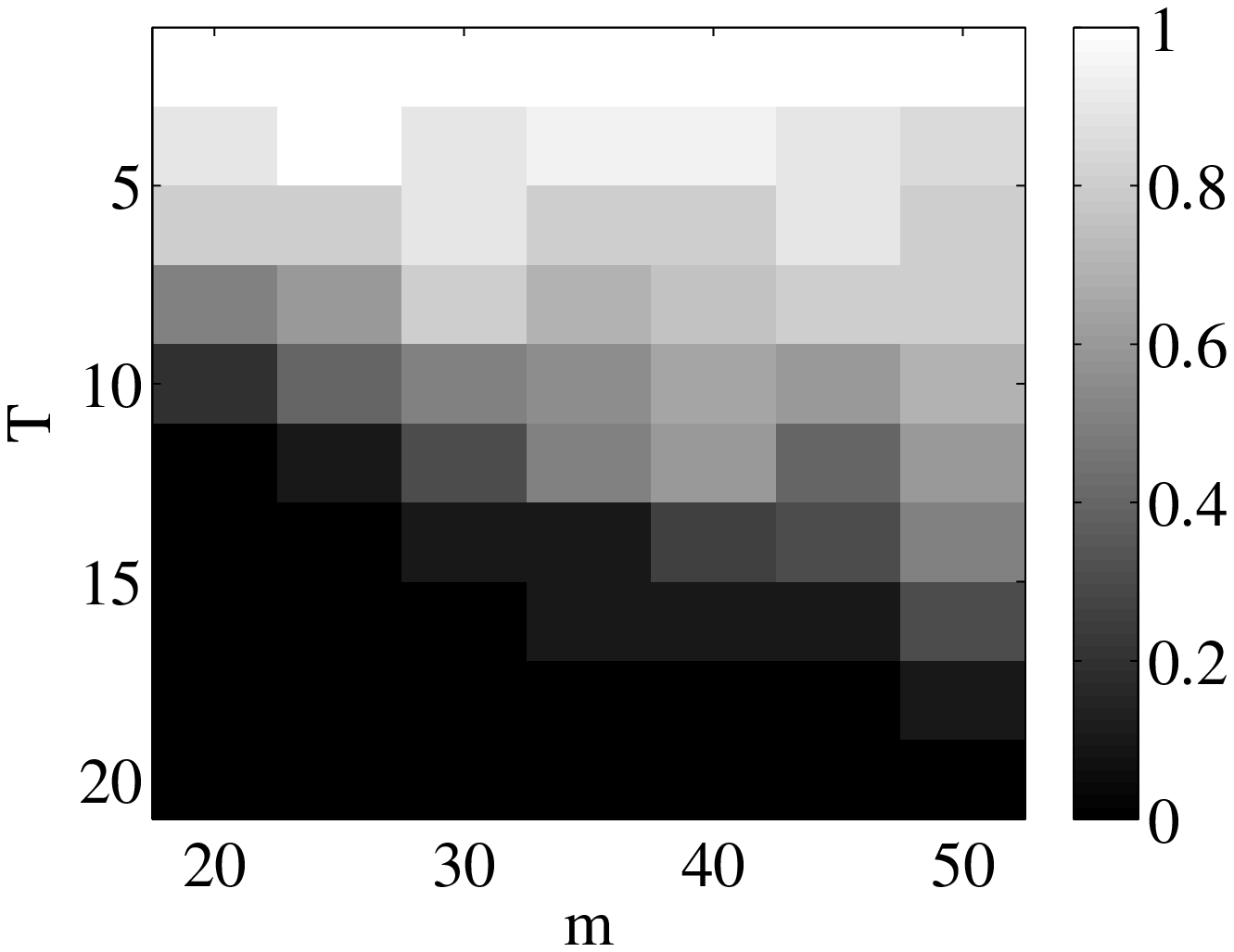}} \label{fig:55a}
}
\subfigure [] {
\resizebox{4cm}{!}{
\includegraphics{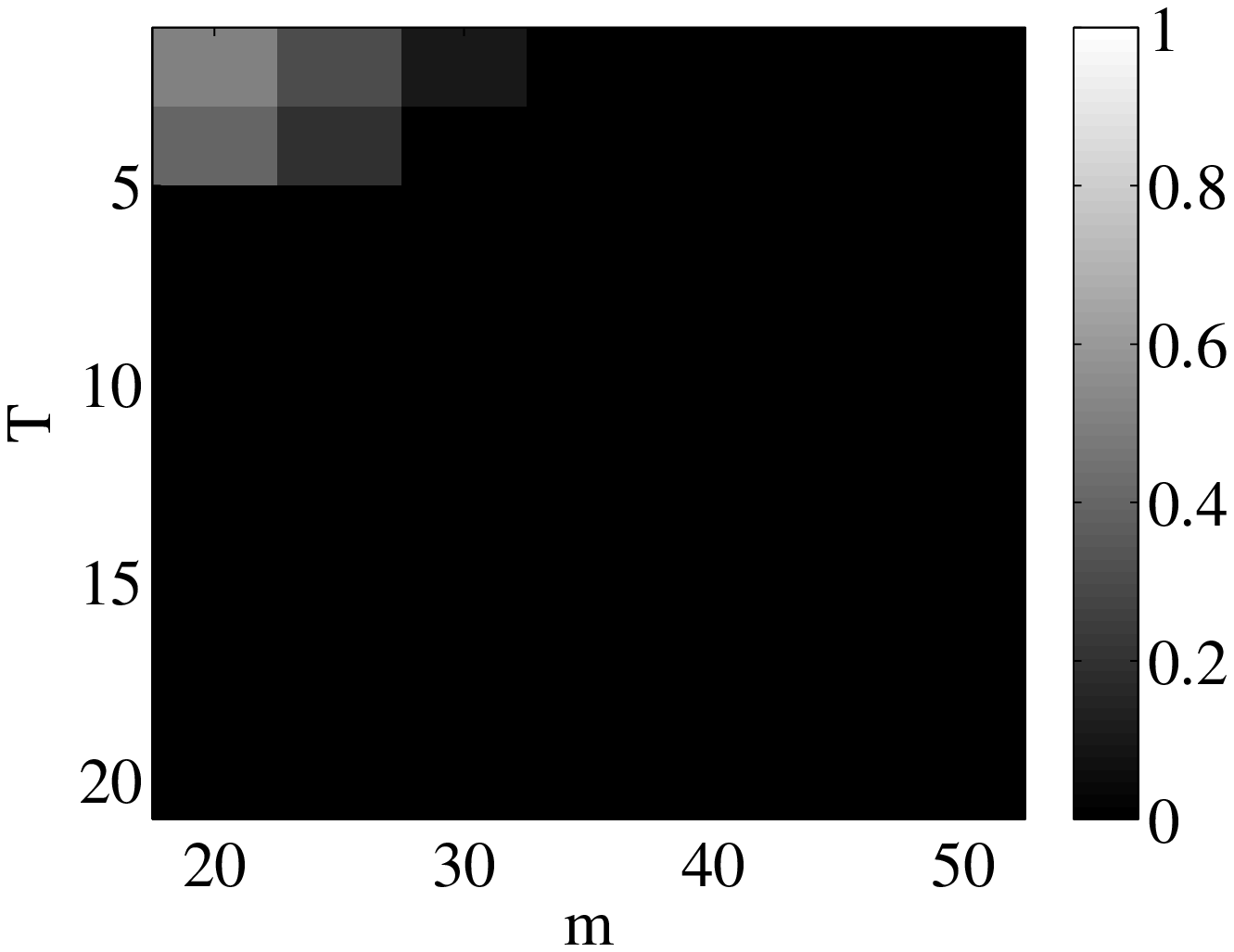}} \label{fig:55b}
}
\subfigure [] {
\resizebox{4cm}{!}{
\includegraphics{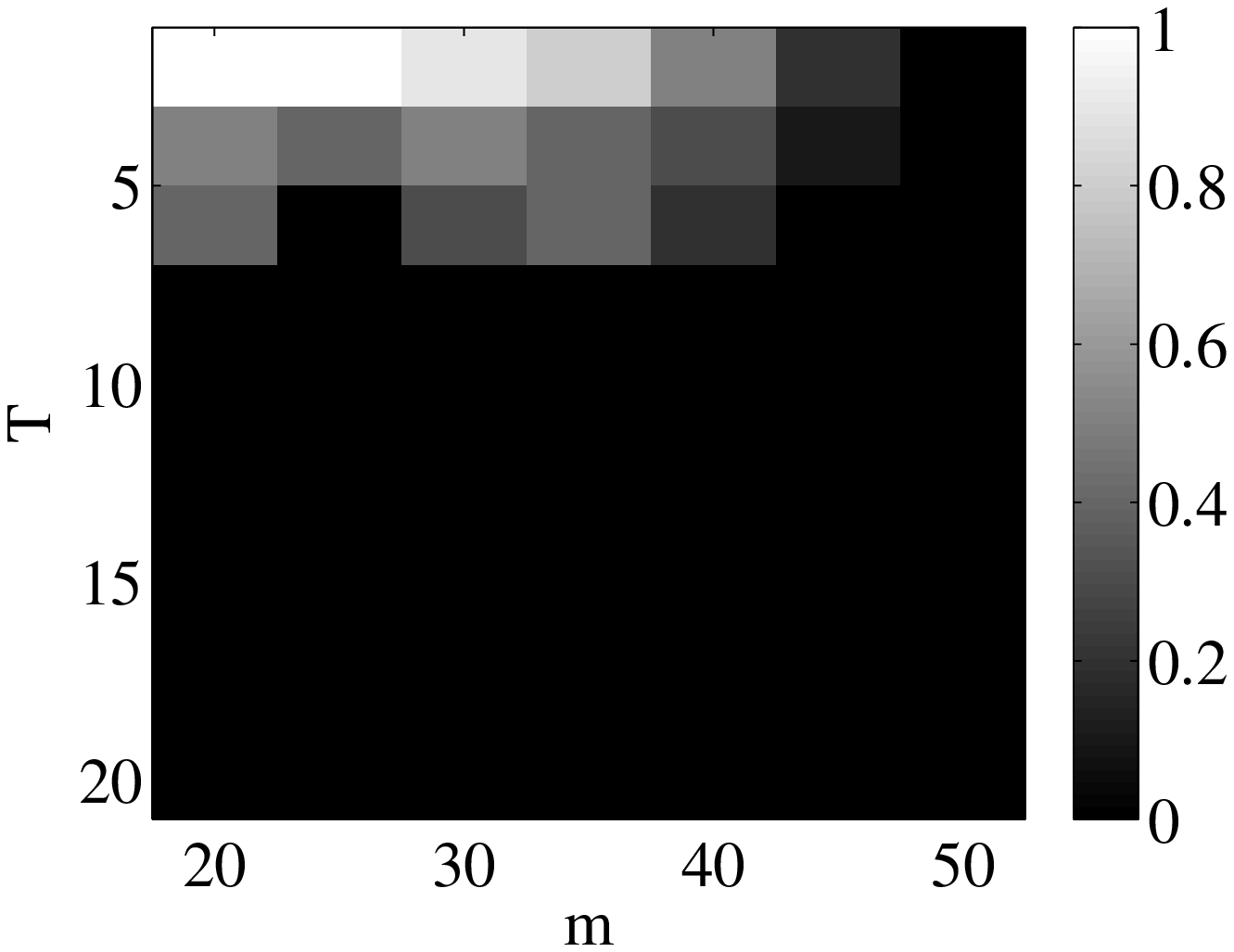}} \label{fig:55c}
}
\caption{Comparison of probability of exact rank recovery obtained by (a) CMEN, (b) RMDE with cross-validation and (c) RMDE with continuation technique for $N = 50$.}\label{fig:55}
\end{figure}
The white region in Fig.~\ref{fig:55a} and Fig.~\ref{fig:55b} correspond to the probability of exact rank recovery obtained by CMEN and RMDE, respectively. The white area in Fig.~\ref{fig:55a} is wider than the white areas in Fig.~\ref{fig:55b} and Fig.~\ref{fig:55c} which means that CMEN is more successful in exact rank recovery compare to RMDE. This is due to the fact that in RMDE, $\eta^*$ is obtained based on the generalization performance (minimum test error) which does not necessarily guarantees exact rank recovery.
Moreover, in CMENA we use a quadratic bound on the main objective function which results in a closed-form expression for the proximal operator.  Based on Eckart-Young \cite{stewart1993early} a low-rank matrix has the lower error in terms of quadratic cost function. Another observation is that the white area in RMDE with continuation technique is slightly wider than RMDE with cross-validation technique. This could be due to the fact that in the continuation technique we start with a very low-rank matrix $\Lambda$ and increase the rank gradually until we reach a targeted value, whereas in the cross-validation technique we keep the regularization parameter constant throughout the optimization.

For $N = 500$, we scan the number of features and rank of matrix $\Lambda$ over a grid of $(m,T)$ with $m$ ranging through $19$ equispaced points in the interval $[100,1000]$ and $T$ ranging through $20$ equispaced points in the interval $[5,100]$. Due to the high computational complexity of scanning through different values of $\eta$ in RMDE with cross-validation, and better result in terms of exact rank recovery in RMDE with continuation technique on small scale data ($N=50$), we compare rank recovery between CMEN and RMDE with continuation technique in this case. Figure~\ref{fig:56a} and \ref{fig:56b} show the phase diagram results for exact rank recovery with CMEN and RMDE (continuation technique) for $N = 500$. We observe that the white area in CMEN approach is wider than the white areas in RMDE approach (better performance in terms of exact rank recovery for CMEN compare to RMDE).
\begin{figure}[ht]
\centering
\subfigure [] {
\resizebox{4cm}{!}{
\includegraphics{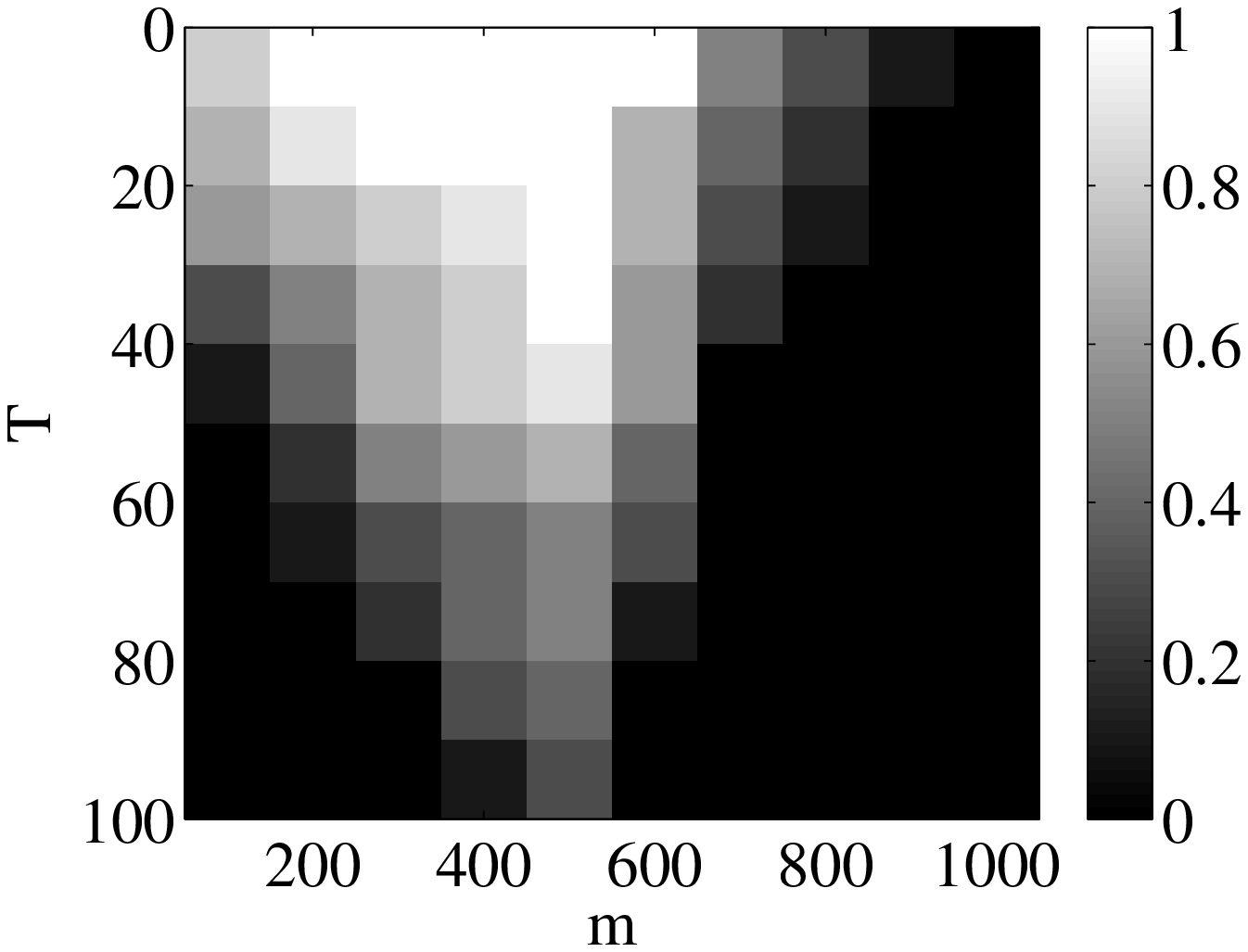}} \label{fig:56a}
}
\subfigure [] {
\resizebox{4cm}{!}{
\includegraphics{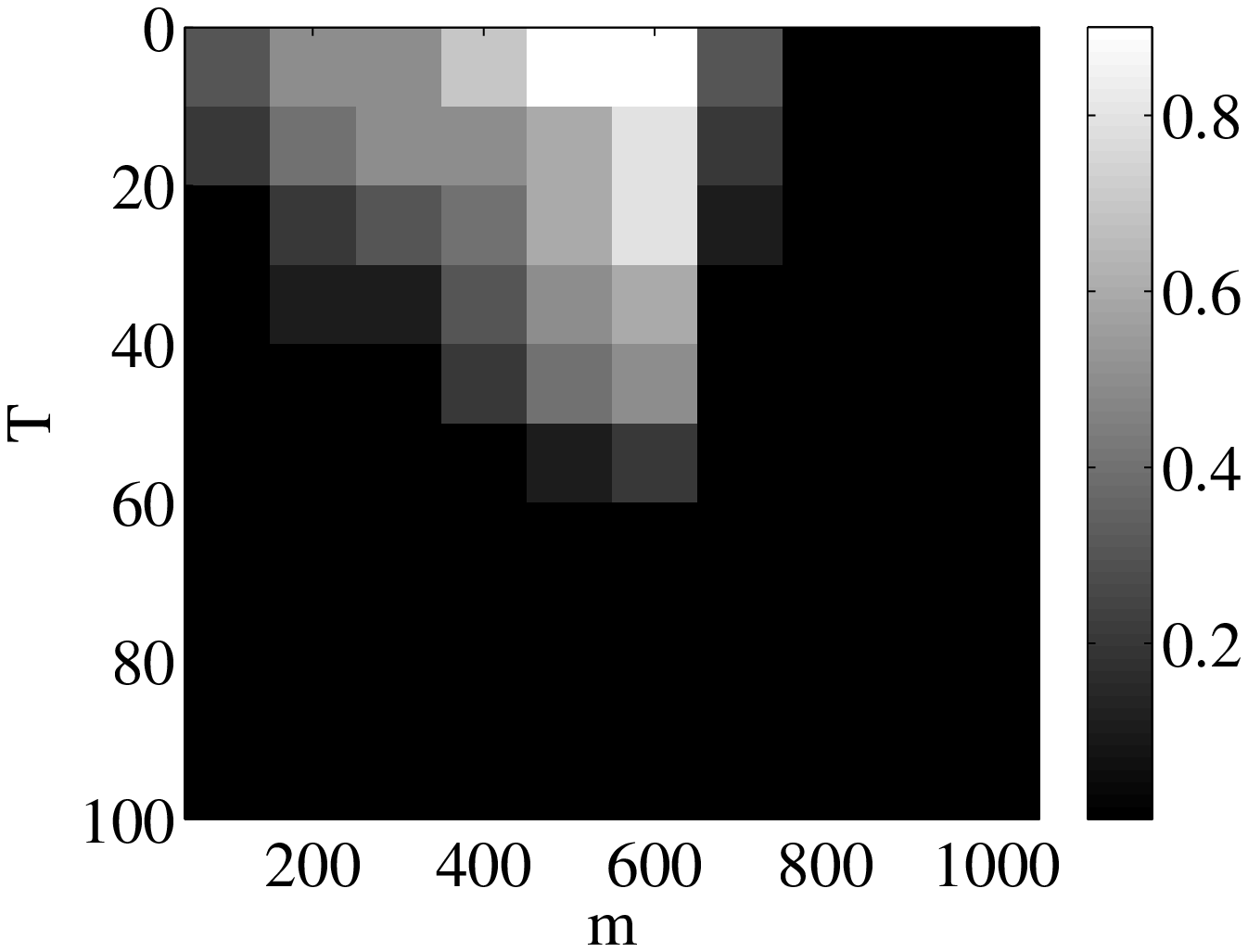}} \label{fig:56b}
}

\caption{Comparison of probability of exact rank recovery obtained by (a) CMEN and (b) RMDE with continuation technique for $N = 500$.}\label{fig:56}
\end{figure}
\subsection{Parameter estimation error}
We compare the test error vs. runtime for both CMEN and RMDE on a synthetic dataset. We construct a low-rank matrix $\Lambda$ and generate $i.i.d.$ samples from the low-rank distribution and estimate matrix $\hat\Lambda$ using maximum likelihood estimation. Then we obtain matrix $\Lambda^*$ using CMEN and RMDE. We consider $N=50$, $m = 100$, $T = 5$, and $T=20$. We randomly choose $70\%$ of the data as a training set and test on the rest of the data over $10$ different realizations. The test error is evaluated as $\sum_i Z(\lambda_i) - \lambda_i^TE_{\hat{p}}[\phi]$, where $i$ indexes all bags in the test set. Figure~\ref{fig:58} shows the results of test error vs. runtime \footnote{We run all algorithms on a standard desktop computer with $2.5$ GHz CPU (dual core) and $4$ GB of memory implemented in MATLAB.}. Figure~\ref{fig:58a} shows the result for $T=5$. Since initially finding the true model with correct rank in CMEN is computationally expensive (due to dual variable update), we observe that RMDE has lower generalization test error than CMEN in the beginning. However, we observe that overall the generalization test error in CMEN decreases faster than RMDE. In Fig.~\ref{fig:58b}, the result is shown for $T=20$. We see that by increasing the complexity of the model, it takes longer for CMEN and RMDE to find the correct model.
\begin{figure}[ht]
\centering
\subfigure [] {
\resizebox{4cm}{!}{
\includegraphics{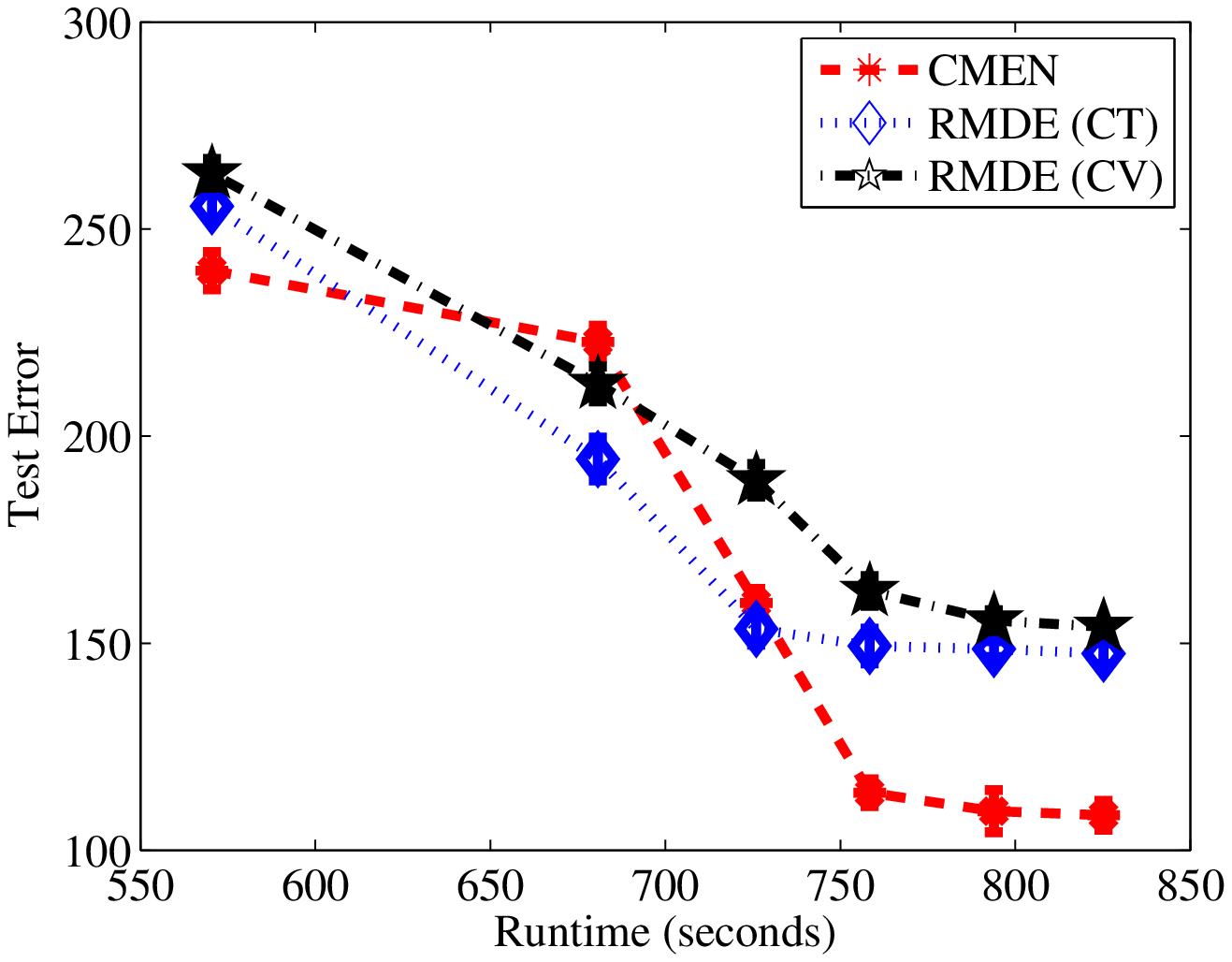}} \label{fig:58a}
}
\subfigure [] {
\resizebox{4cm}{!}{
\includegraphics{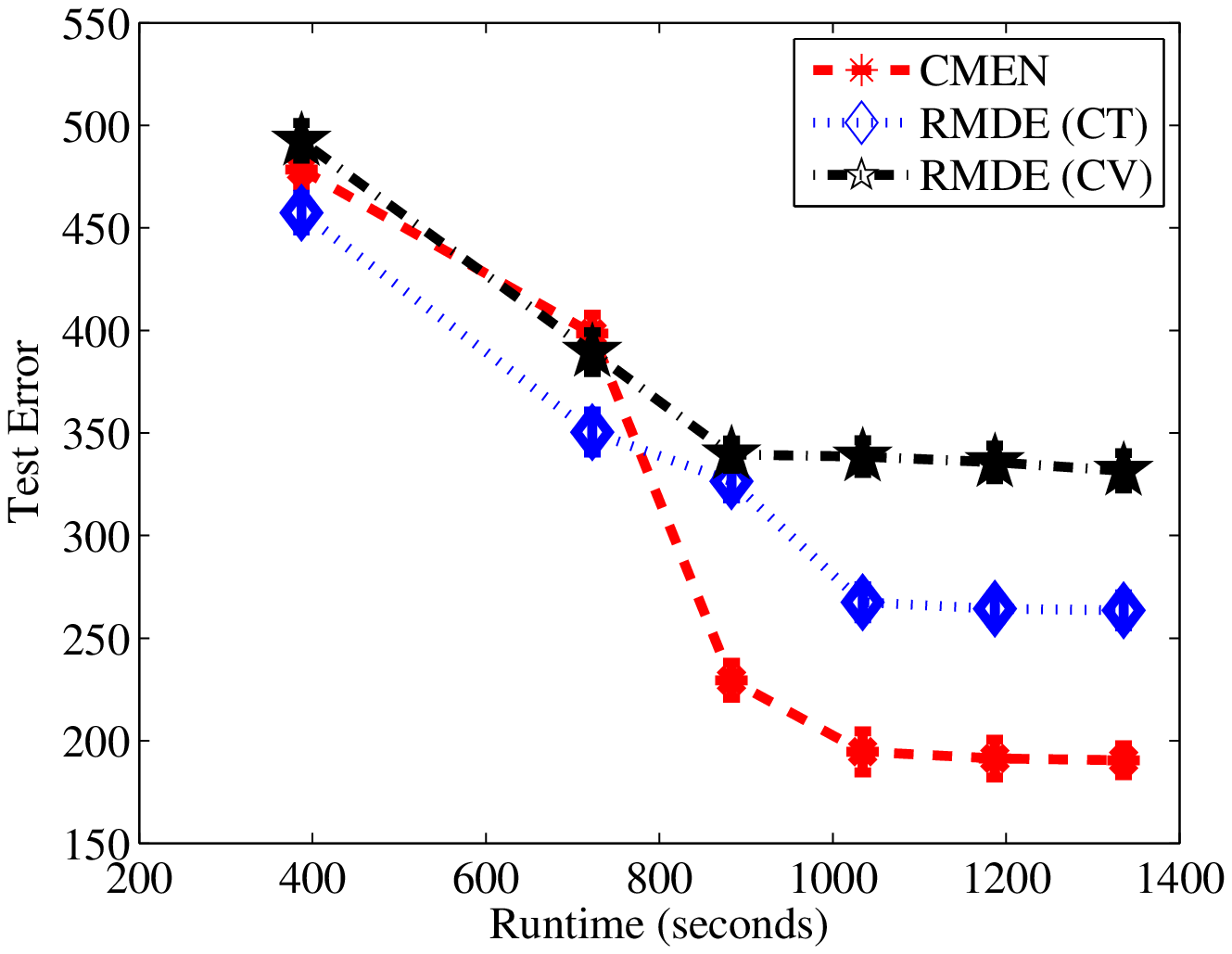}} \label{fig:58b}
}
\caption{Comparison of generalization test error ($\sum_i Z(\lambda_i) - \lambda_i^TE_{\hat{p}}[\phi]$) vs. runtime for $N=50$, $m=100$, and (a) $T=5$, (b) $T=20$.}\label{fig:58}
\end{figure}
\subsection{Dimension reduction}
The purpose of this section is to illustrate how dimension reduction can be achieved using the $\psi$ obtained by CME. Since all the datasets are high dimensional, we use PCA as a preprocessing step. Figure~\ref{fig:Intuition_PCA} depicts the whole process of implementing our approach for one image in the Corel1000 dataset \cite{duygulu2006object}. We use the block representation of the image followed by PCA to reduce the dimension. The image is represented as a bag of instances where each instance corresponds to a small rectangular patch of pixels. The feature vector describing each patch is the raw pixel intensities (RGB) with PCA applied to reduce the dimension. We perform the CMEN approach to learn a p.d.f.~over the block representation of the image.

After performing the nuclear norm minimization in (\ref{eq:CCMaxEnt24}) on the Corel1000 dataset, we select one image as an example. Then, we choose the first few bases of matrix $\psi$ obtained by (\ref{eq:CCMaxEntpsi}) to represent the image as a linear combination of these basis functions. Figure \ref{fig:PCA} shows that the contour plots of these basis functions. To provide intuitive understanding, we name each basis $\psi$ following the content of the image corresponding to instances near the peaks of $\psi$ (concentration of data points). The first column of Fig.~\ref{fig:PCA} is an image and its corresponding estimated density. The other columns show each $\psi_i$ and the part of the image that corresponds to that $\psi_i$.

\begin{figure}[!ht]
\centering
\includegraphics[scale = 0.4]{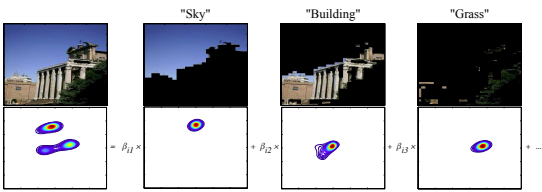}
\caption{Dimension reduction in the space of the distribution obtained by the bases $\psi$. The first column shows the image and corresponding density estimation. The other columns show each $\psi$ and part of the image that corresponds to that $\psi$.}
   \label{fig:PCA}
\end{figure}
\subsection{KL-divergence similarity}
For classification and retrieval, it is useful to have a similarity measure between bags. The Kullback-Leibler (KL) divergence between two estimated distributions provides such a similarity measures \cite{viola2006multiple}. The KL divergence between two distributions obtained by the maximum entropy approach has a closed form:
\begin{eqnarray*}
D(p_{\lambda_i}\|p_{\lambda_j}) = (\lambda_i-\lambda_j)^TE_{p_{\lambda_i}}[\Phi] - (Z(\lambda_i)-Z(\lambda_j)).
\end{eqnarray*}
We symmetrize the divergence by adding $D(p_{\lambda_i}\|p_{\lambda_j}) + D(p_{\lambda_j}\|p_{\lambda_i})$.
\begin{eqnarray*}
D(p_{\lambda_i}\|p_{\lambda_j}) + D(p_{\lambda_j}\|p_{\lambda_i}) = (\lambda_i-\lambda_j)^T(E_{p_{\lambda_i}}[\Phi]-E_{p_{\lambda_j}}[\Phi]).
\end{eqnarray*}
Figure~\ref{fig:neighbor} shows a set of images and their nearest images identified by KL-divergence similarity. We observe that by using the KL-divergence similarity, the nearest neighbor images resemble the main images which validates the efficacy of the proposed similarity measure. Figure~\ref{fig:neighbor_fail} shows failure examples in which the nearest neighbor image comes from a different class than the original image. We hypothesis that this is due to the dominance of the color features.
 \begin{figure}[!ht]
\centering
\includegraphics[scale = 0.4]{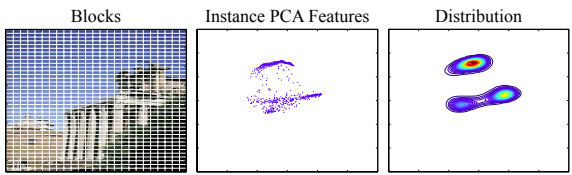}
\caption{The whole ME process from bag representation to fitting a distribution. The figures from left to right shows the following: $(1)$ how an images is represented as a bag of instances (blocks), $(2)$ The $2D$ PCA features of each instance $(3)$ the density fitted to the data using the maximum entropy principle.}
   \label{fig:Intuition_PCA}
\end{figure}
\subsection{Datasets}
We also evaluate the classification accuracy of the proposed KL-divergence based similarity measure when used in distance-based multi-instance algorithms such as Citation-kNN \cite{Wang2000} and bag-level kernel SVM \cite{gärtner2002multi}.
We compare KL-divergence to bag-level distance measures that rely on pairwise instance-level comparisons, namely average Hausdorff distance \cite{Wang2000} and the RBF set kernel \cite{gärtner2002multi}, both in terms of accuracy and runtime.
The comparison is conducted over four datasets, i.e., the Corel1000 image dataset \cite{duygulu2006object} Musk1, Musk2 \cite{dietterich1997solving}, and Flowcytometry \cite{carter2009information}. The Corel1000 \cite{duygulu2006object} image dataset consists of $10$ different classes each containing $100$ images. We use $50$ randomly subsampled images from $4$ classes: \textit{`buildings', `buses', `flowers', and `elephants'}. We represent each image (bag) as a collection of instances, each of which corresponds to a $10\times 10$ pixel block, and is described by a feature vector of all pixel intensities in $3$ color channels (RGB). The Musk1 dataset \cite{dietterich1997solving} describes a set of 92 molecules of which 47 are judged by human expert to be musks and the remaining 45 molecules are judged to be non-musk. The Musk2 dataset \cite{dietterich1997solving} is a set of 102 molecules of which 39 are judged by human experts to be musks and the remaining 63 molecules are judged to be non-musks. Each instance corresponds to a possible configuration of a molecule. The Flowcytometry dataset consists of $5d$ vector reading of multiple blood cell samples for each one of $43$ patients. For each patient, we have two similar cell characteristics with respect to the antigens surface which are called $1)$ chronic lymphocytic leukemia (CLL) or $2)$ mantle cell lymphoma (MCL). Each patient is associated with one bag of multiple cells (instances). Table~\ref{tb:MaxEnt11} summarizes the properties of each dataset.
\begin{table}
\centering
\caption{Datasets}
  \begin{tabular}{| c  | c | c | c | c |}
    \hline
Dataset  & bags & no. of class & Ave. inst/bag & dim\\ \hline
Corel1000 (4class) & $200$ & $4$ & $950$ & $300$\\\hline
Musk1 & $92$ & $2$ & $4.5$ & $166$\\\hline
Musk2 & $102$ & $2$ & $64.7$ & $166$\\\hline
Flowcytometry & $43$ & $2$ & $5664$ & $5$\\\hline
  \end{tabular}
\label{tb:MaxEnt11}
\end{table}

\subsection{Experimental setup}\label{sec:experimental-setup}
We use classification accuracy as an evaluation metric. In all experiments, we use the preprocessed datasets obtained by PCA. We perform $10$-fold cross-validation over all datasets. As baselines, we implement a modified version of Citation-kNN \cite{Wang2000} replacing the Hausdorff distance with KL-divergence, and a bag-level SVM with the kernel for two bags $X$ and $X'$ defined as $K(X,X') = e^{-\gamma D_{KL}(X,X')}$,  $K(X,X') = e^{-\gamma D_{Haus}(X,X')}$, and the RBF set kernel used by \cite{gärtner2002multi}. Below we state the ranges of all tunning parameters for these algorithms used in our experiments. We compared CMEN with RMDE with cross-validation and RMDE with continuation technique. We use a grid of $\{10^{-4},10^{-3},\ldots,10^3,10^4\}$ for the regularization parameter $\eta$. All of the datasets use features with dimension reduced by PCA. We use a grid of  $\{2, 3, 4, 5, 6, 7\}$ for the feature dimension after applying PCA. The Citation-kNN algorithm has two parameters- the number of nearest neighbors $k$, and the number of ``citers'' $k'$. We use a grid of $\{1, 5, 10, 15, 20\}$ for k and $\{5, 10, 15, 20, 25\}$ for $k'$. The SVM has two  parameters- the bandwith of RBF kernel $\gamma$, and the penalty factor $C$. We use a grid of $\{2^{-9}, 2^{-8}, \ldots, 2^{0}\}$ for $\gamma$, and a grid of $\{2^{0}, 2^{1}, \ldots, 2^{9}\}$ for C. For the basis functions used in constructing the maximum entropy distribution space, we propose $\phi_{2k} = \cos (g_k^T x)$ and $\phi_{2k-1}=\sin (g_k^T x)$, where $g_k \sim {\cal N} (0, I)$ i.i.d for $k=1,2,\ldots,m/2$. In \cite{rahimi2007random}, a similar transformation is used to approximate Gaussian kernels.
\begin{figure}[!ht]
\centering
\includegraphics[scale = 0.2]{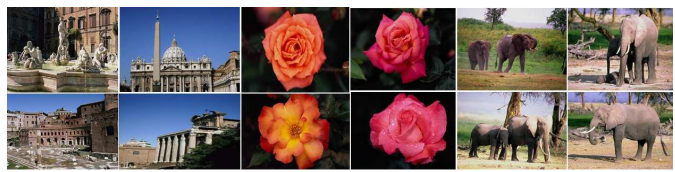}
\caption{Top: Query image. Bottom: Nearest-neighbor based on KL-divergence. This figure shows success in retrieving the corresponding image.}
   \label{fig:neighbor}
\end{figure}
\begin{figure}[!ht]
\centering
\includegraphics[scale = 0.15]{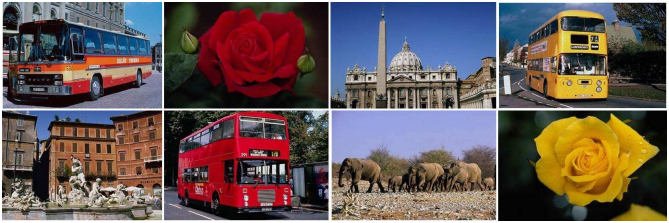}
\caption{Top: Query image. Bottom: Nearest-neighbor based on KL-divergence.  This figure shows failure in retrieving the corresponding image.}
   \label{fig:neighbor_fail}
\end{figure}
\subsection{Classification }
The results of classification accuracy for citation-kNN for four datasets are shown in Fig.~\ref{fig:CCMaxEntcomparison}. We compared the classification accuracy with Citation-kNN using KL-divergence and Hausdorff distance. The KL divergence is computed from $3$ different distribution estimates: $1)$ RMDE (CV): RMDE with cross-validation, $2)$ RMDE (CT): RMDE with continuation technique, and $3)$ CMEN. We observe that CMEN has slightly better classification performance than RMDE in musk and image datasets, where in Flowcytometry dataset RMDE (CT) is performing better. However, the difference is not very significant.
 \begin{figure}[!ht]
   \begin{center}
\subfigure[Citation-kNN, Corel1000] {\label{fig:comparison1}
\resizebox{4cm}{!}{
\includegraphics{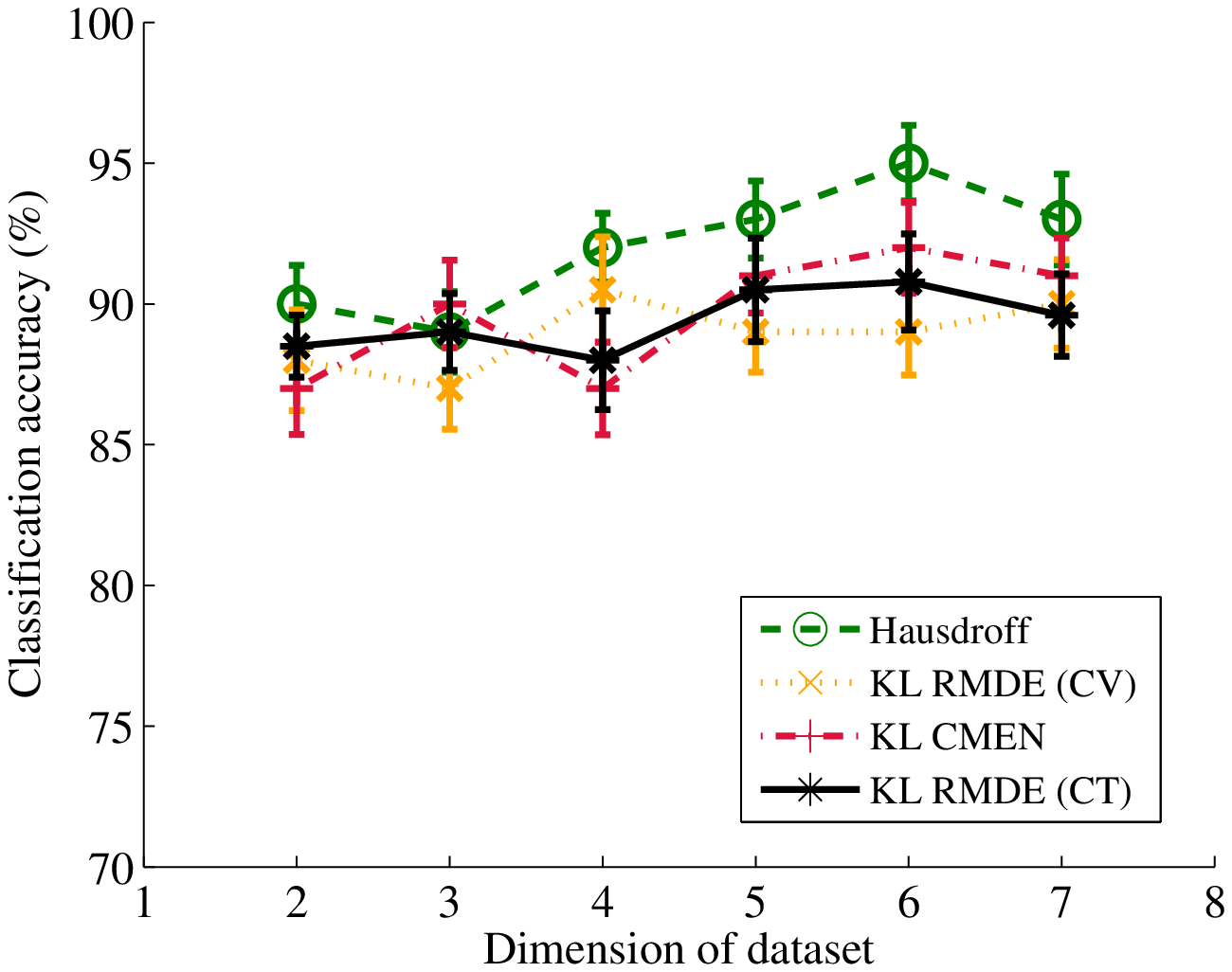}}

}
\subfigure[Citation-kNN, Musk1] {\label{fig:comparison4}
\resizebox{4cm}{!}{
\includegraphics{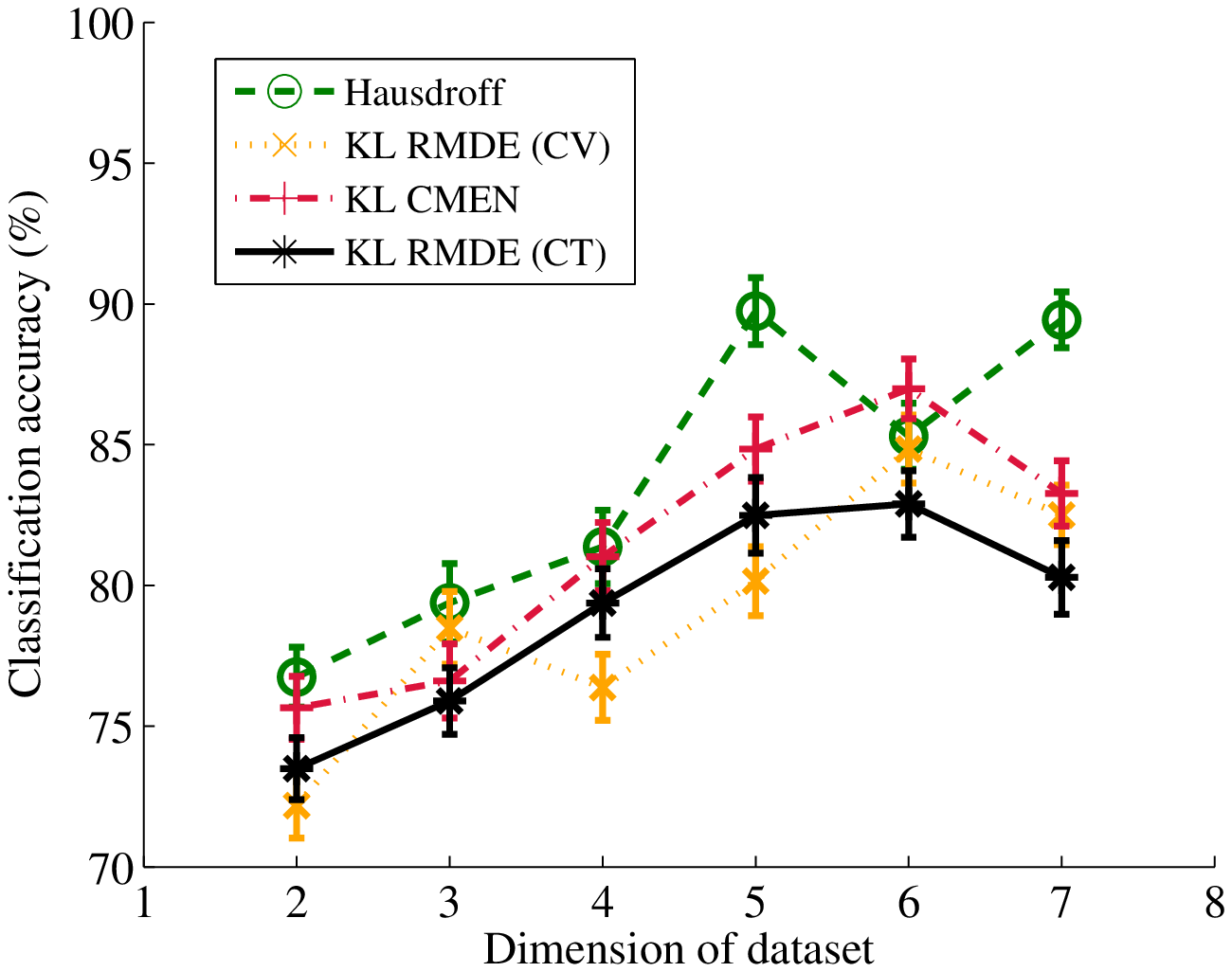}}

}
\subfigure[Citation-kNN, Musk2] {\label{fig:comparison2}
\resizebox{4cm}{!}{
\includegraphics{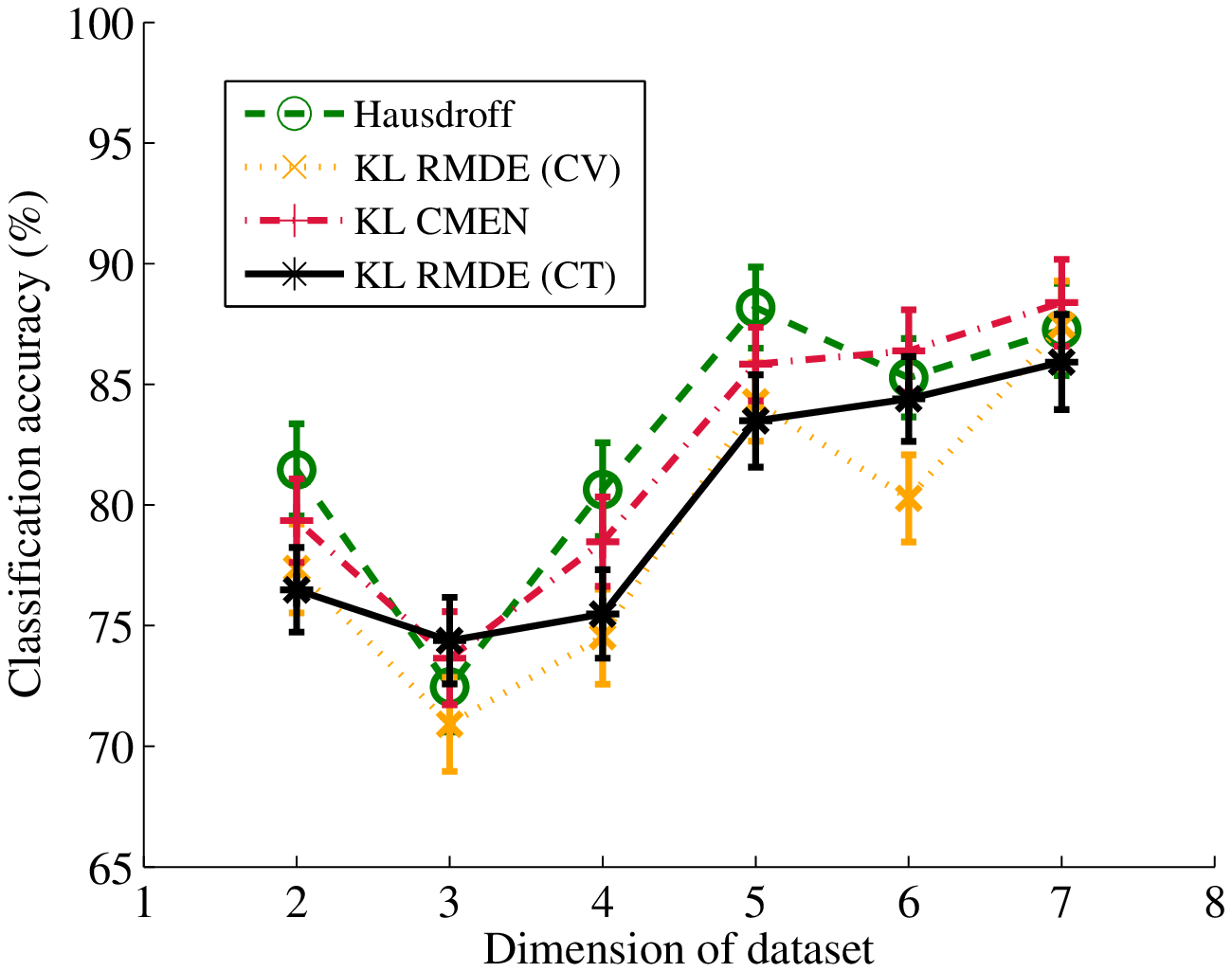}}

}
\subfigure[Citation-kNN, Flowcytometry] {\label{fig:comparison5}
\resizebox{4cm}{!}{
\includegraphics{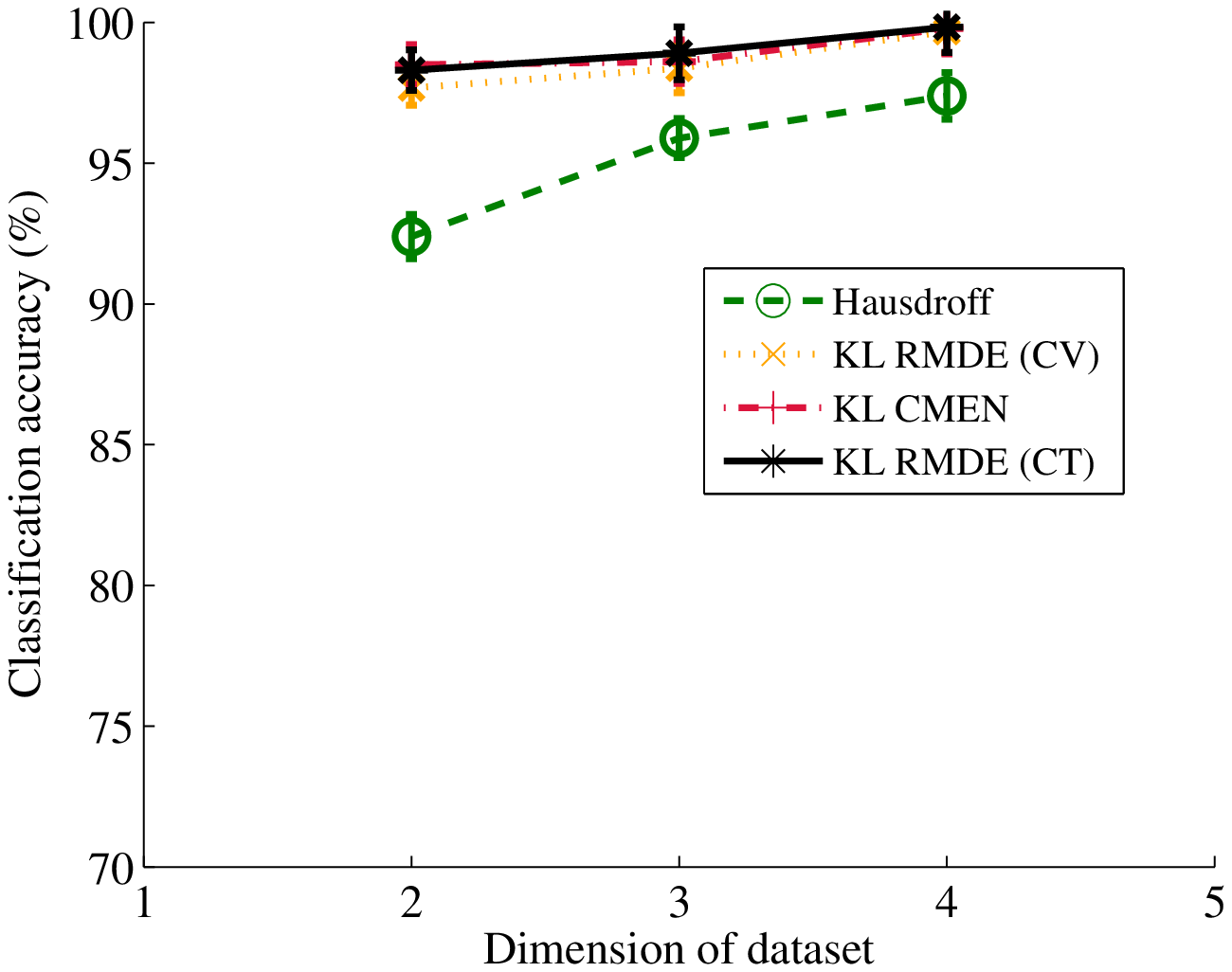}}

}

   \end{center}
   \caption{ Classification accuracy results for (a) Corel1000, (b) Musk1  (c) Musk2 and (d) Flowcytometry.}
   \label{fig:CCMaxEntcomparison}
\end{figure}
Figure~\ref{fig:CCMaxEntcomparison2} shows the results 
for bag-level SVM with the RBF set kernel, the average Hausdorff distance kernel, and the KL divergence kernel obtained by RMDE and CMEN. KL divergence has better classification performance than Hausdorff distance in musk1, musk2, and flowcytomery datasets. In Corel1000 image dataset, Hausdorff distance is performing better than KL divergence. Accuracy results are very close to all methods using KL divergence.

 \begin{figure}[!ht]
   \begin{center}
\subfigure[SVM, Corel1000] {\label{fig:156}
\resizebox{4cm}{!}{
\includegraphics{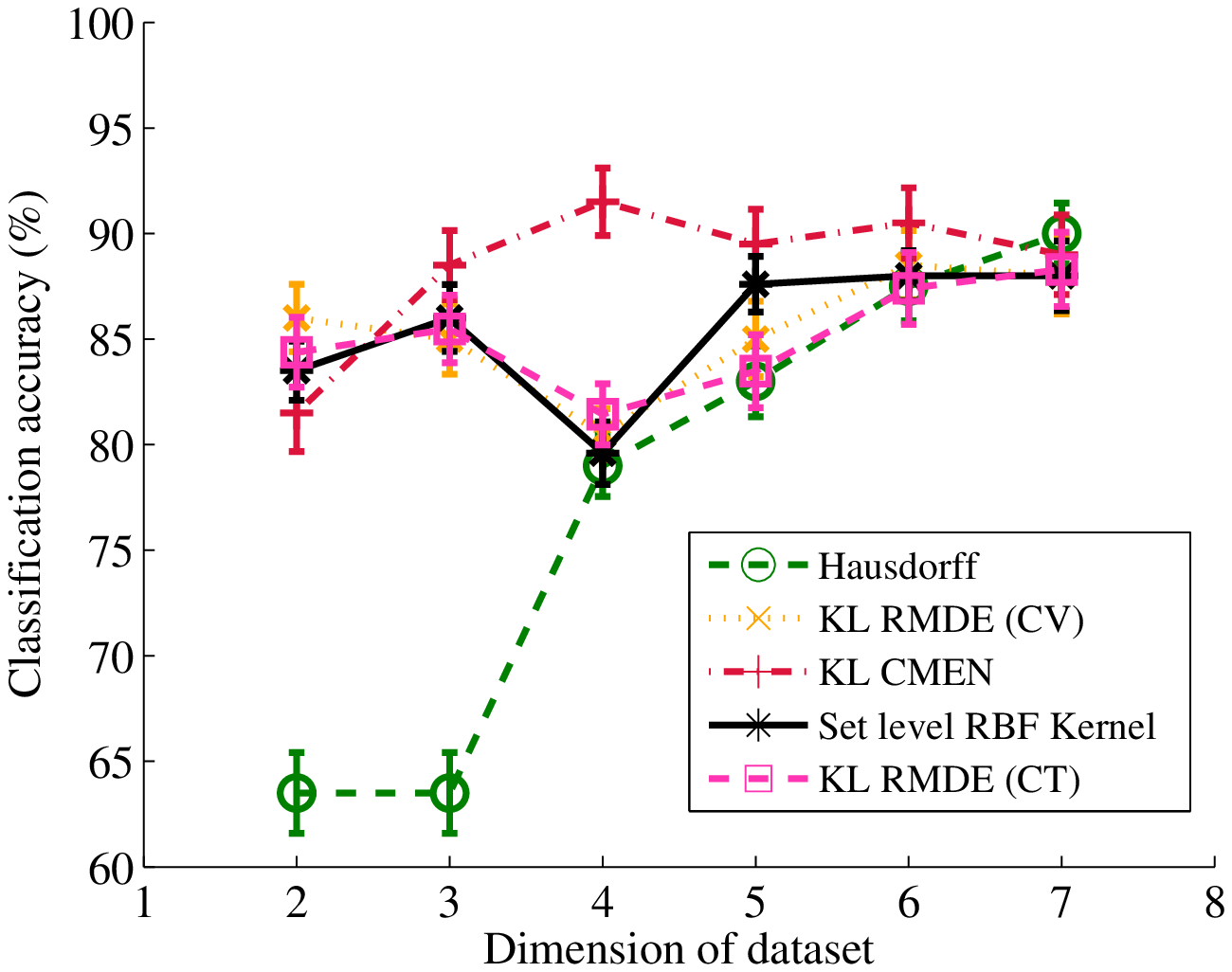}}
}
\subfigure[SVM, Musk1] {\label{fig:456}
\resizebox{4cm}{!}{
\includegraphics{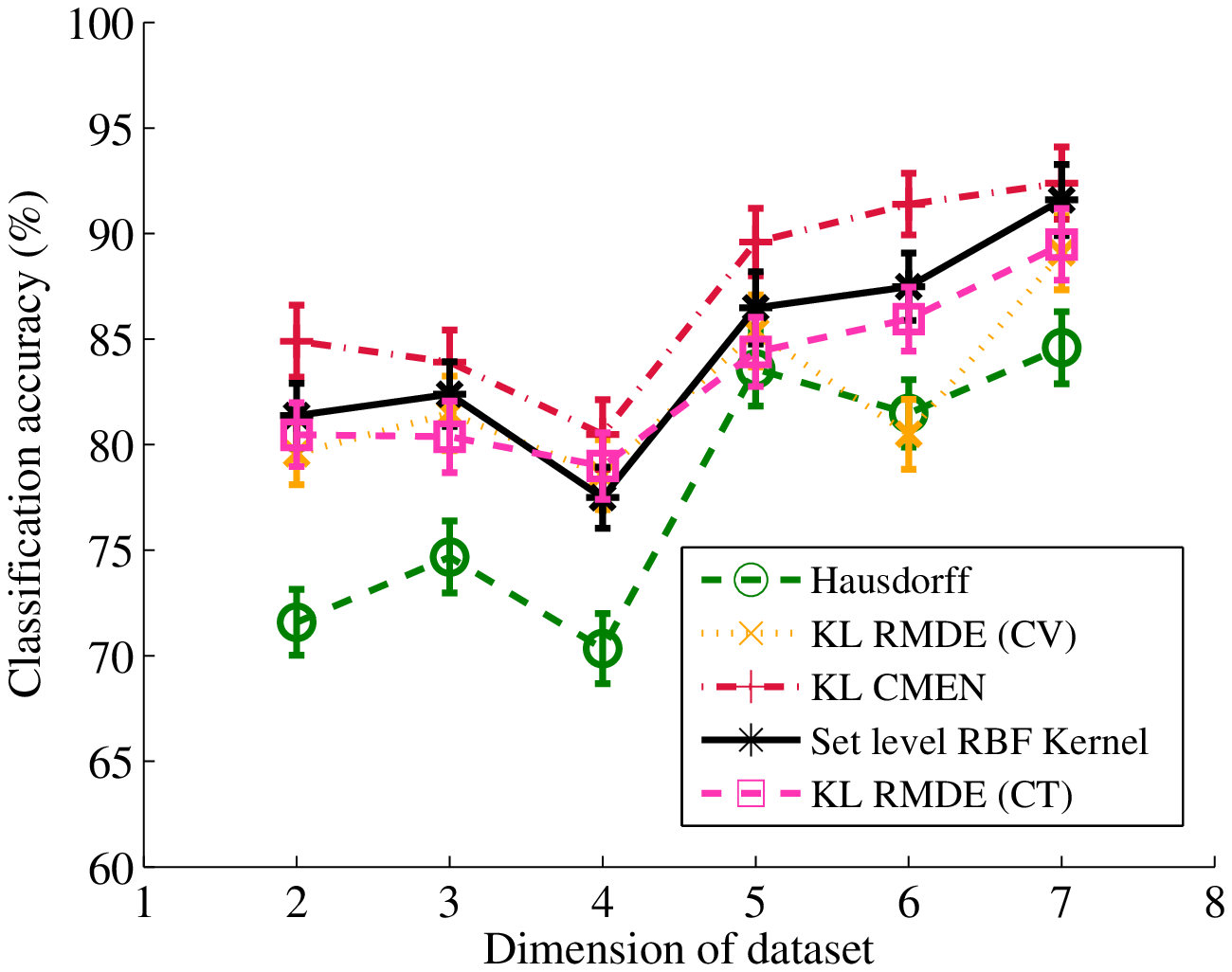}}

}
\subfigure[SVM, Musk2] {\label{fig:256}
\resizebox{4cm}{!}{
\includegraphics{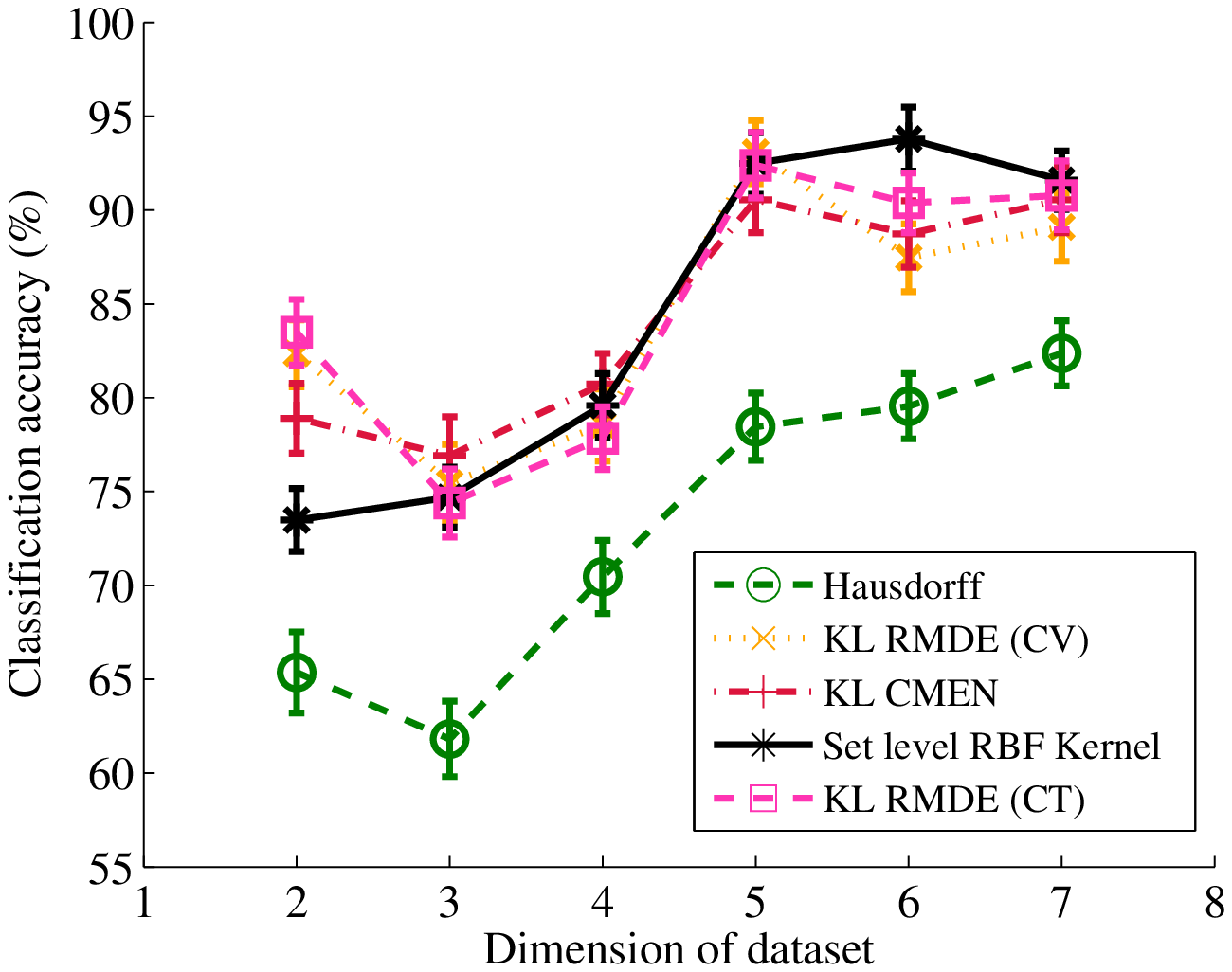}}

}
\subfigure[SVM, Flowcytometry] {\label{fig:556}
\resizebox{4cm}{!}{
\includegraphics{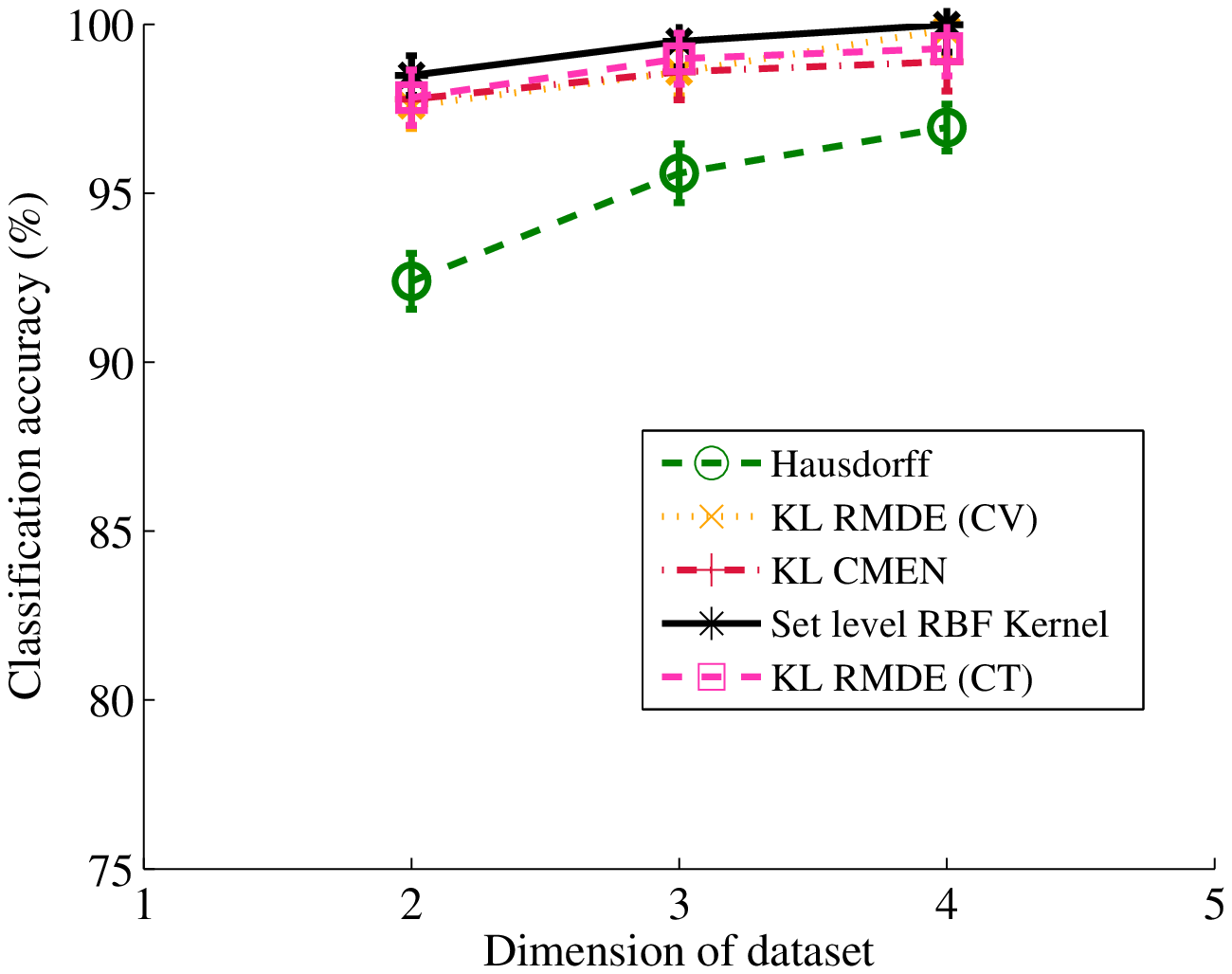}}

}
\psfrag{KL CMENA}{KL CMEN}

   \end{center}
   \caption{ Classification accuracy results for (a) Corel1000, (b) Musk2 (c) Musk1 and (d) Flowcytometry. Set level RBF kernel accuracy is provided for reference.}
   \label{fig:CCMaxEntcomparison2}
\end{figure}
\subsection{Comparison CMEN vs. Kernel Density Estimation}
{We compare CMEN to a nonparametric kernel density estimation (KDE) in classification application for two real world datasets, namely: Corel1000 and  Musk1. The KDE of $p_i(x)$ is defined as 
\begin{eqnarray}
{p}_i(x)=\frac{1}{n_i \cdot h}\sum_{j=1}^{n_i}K\left(\frac{x-x_j}{h}\right),
\end{eqnarray}
where $n_i$ is the number of sample in bag $i$, $K$ is a kernel function, and $h$ is the bandwidth. We select the Gaussian kernel $K(x)=\frac{1}{\sqrt{|2\pi\Sigma|}}\exp\left(-\frac{1}{2}x^T\Sigma^{-1}x\right)$ where $\Sigma$ is a covariance matrix. We use the maximal smoothing principle \cite{terrell1990maximal} for determining the bandwidth parameter $h$ and $\Sigma=I$. We use the two mentioned classification techniques, citation k-NN and bag-level SVM to compare the accuracy performance of KDE vs. CMEN. For both approaches, we use KL-divergence as a similarity measure. All experimental setup is as described in Section~\ref{sec:experimental-setup}.}
\begin{figure}[!ht]
   \begin{center}
\subfigure[Citation-kNN, Corel1000] {\label{fig:1}
\resizebox{4cm}{!}{
\includegraphics{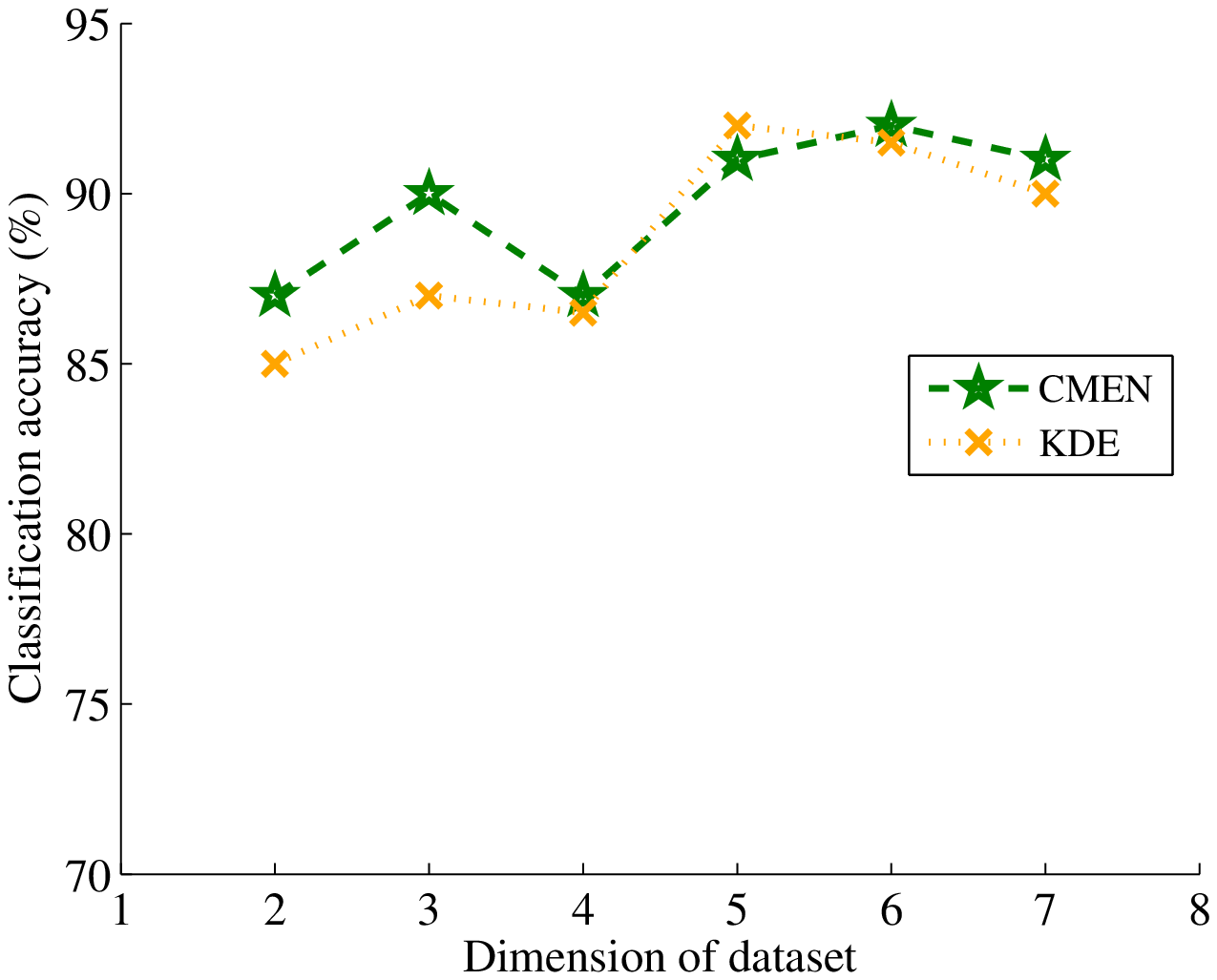}}

}
\subfigure[Citation-kNN, Musk1] {\label{fig:4}
\resizebox{4cm}{!}{
\includegraphics{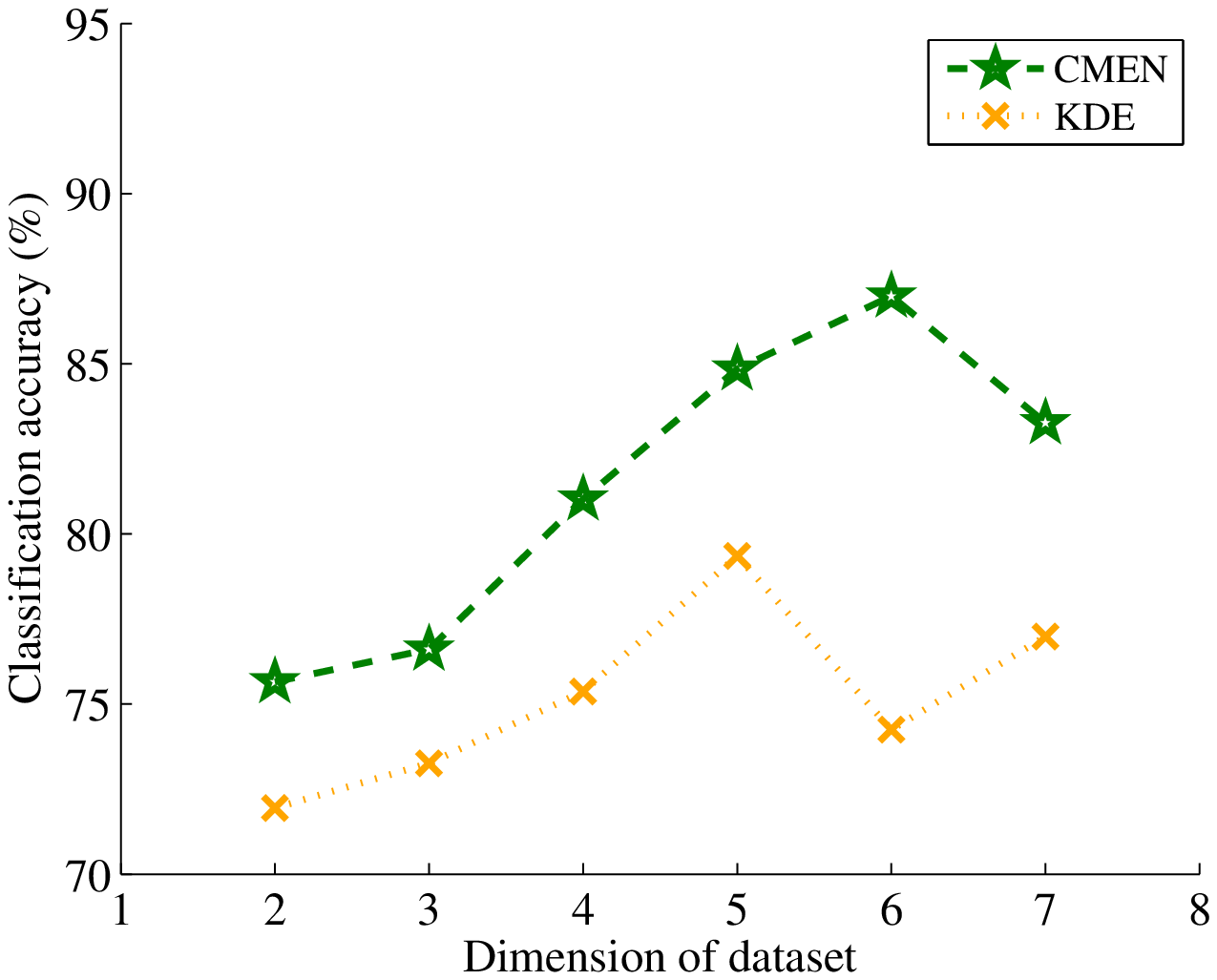}}

}
\subfigure[SVM, Corel1000] {\label{fig:2}
\resizebox{4cm}{!}{
\includegraphics{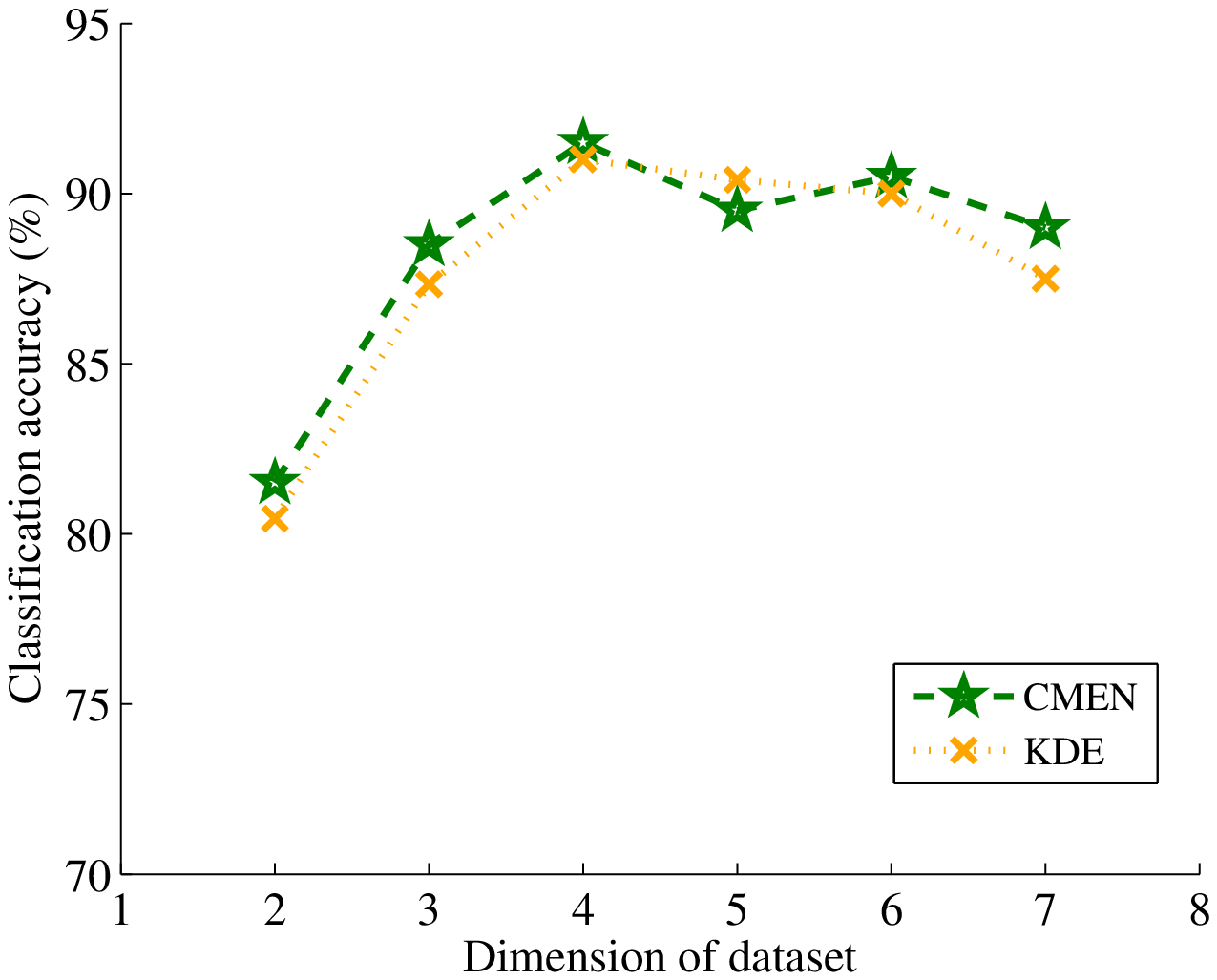}}
}
\subfigure[SVM, Musk1] {\label{fig:5}
\resizebox{4cm}{!}{
\includegraphics{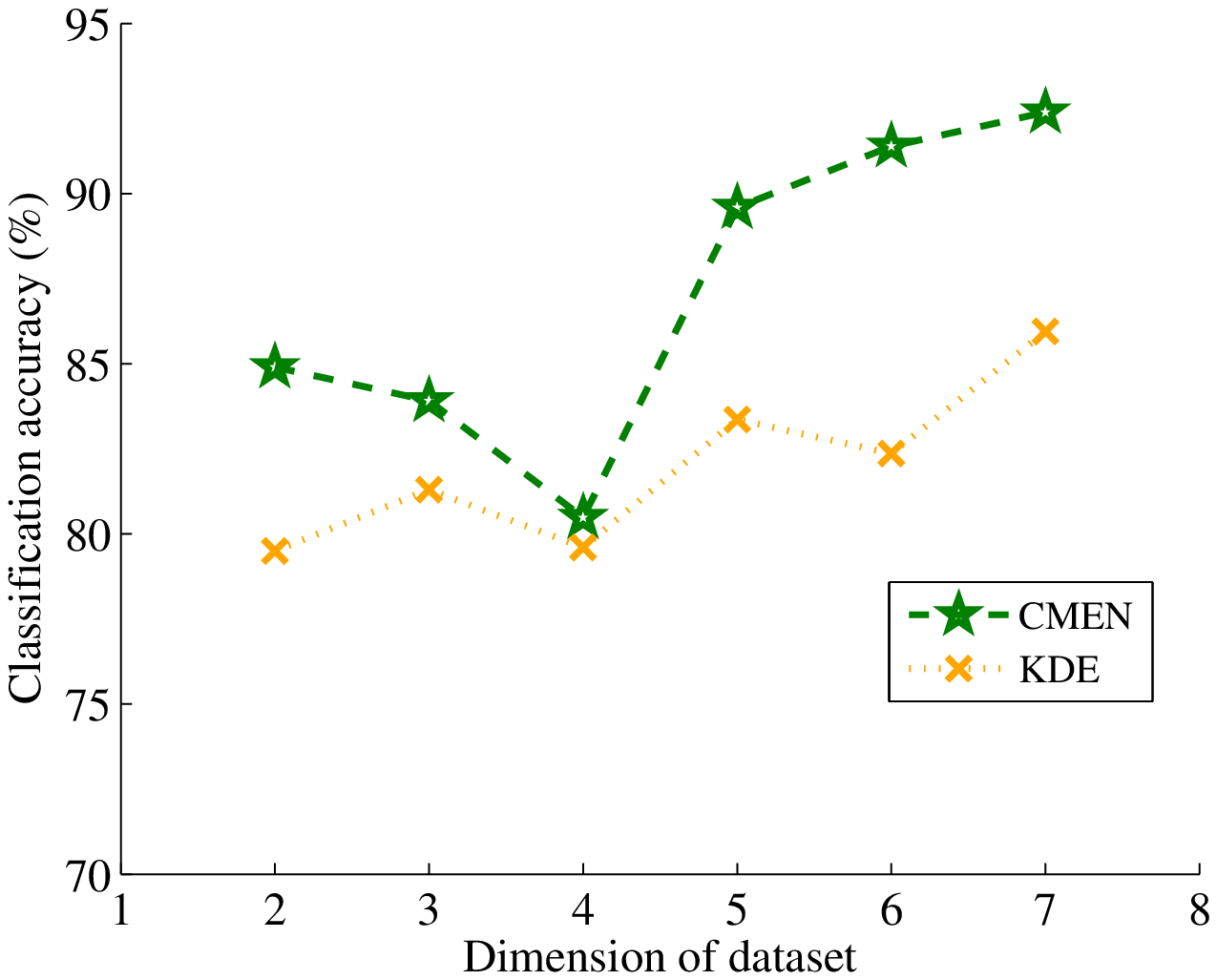}}
}

   \end{center}
   \caption{Comparison between CMEN and KDE in terms of classification accuracy results.}
   \label{fig:comparison-cmen-kde}
\end{figure}
{Figure~\ref{fig:comparison-cmen-kde} shows the classification accuracy for both CMEN and KDE. The results are comparable for image Corel1000 dataset, but for Musk1 dataset CMEN outperforms KDE in terms of prediction accuracy (maximum accuracy value for CMEN=$93\%$ where for KDE=$86\%$). The reason that CMEN performs better compare to KDE for Musk1 dataset is due to the lack of enough samples in each bag. Lack of enough samples in each bag causes the individual density estimated by KDE to be inaccurate and therefore, the similarity metric used in the classification algorithms is not discriminative. But, CMEN uses the shared structure of the data by enforcing the low-rank structure on parameter $\Lambda$. Standard deviation for both approaches is equal and is in order of $4\%$.}

\subsection{Runtime}
To compare the computational complexity of our algorithm with standard MIL algorithms, we run Citation-kNN and MI-SVM using the MIL toolkit\footnote{http://www.cs.cmu.edu/~juny/MILL/} on the Corel1000 image dataset for different numbers of instances in each bag. To evaluate how the runtime of each algorithm depends on the number of instances in the dataset, we randomly sample varying number of instance from each bag. In Fig.~\ref{fig:time1}, the $x$-axis shows the number of samples in each bag and the $y$-axis shows the elapsed CPU time in seconds. We compare the time complexity of standard MIL algorithm with RMDE (CV), RMDE (CT), and CMENA. The runtime of Citation-kNN and SVM is significantly longer than RMDE and CMENA by several orders of magnitude. Hence our proposed approach achieves superior runtime and similar accuracy to two standard MIL algorithms.
\begin{figure}[!ht]

\centering
\includegraphics[scale = 0.4]{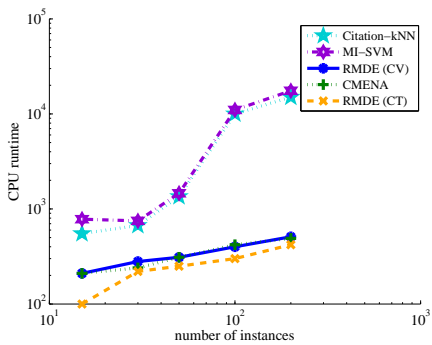}


\caption{Time comparison among Citation-kNN, MI-SVM, RMED, and CMENA.}
   \label{fig:time1}
\end{figure}
The computational complexity of RMDE and CMENA during training is $\mathcal{O}(Nndm)$, where $n$ is average number of instance per bag, $d$ is the dimension of instances, and $m$ is the number of basis functions $\phi$ and during test is $\mathcal{O}(Nm)$. The computational complexity of Hausdorff distance during test is $\mathcal{O}(n^2 N)$. The Hausdorff distance based approach requires no training.

\subsection{Discussion}
The RMDE and CMEN approaches for MIL are significantly faster than other algorithms when there are a large number of instances in each bag. RMDE and CMEN achieve this speedup by summarizing the instances in each bag, thereby avoiding instance-level processing in later steps.
\section{Conclusion}\label{se:5}
In this paper, we proposed a confidence-constrained maximum entropy approach for learning from multi-instance data. The proposed approach used the idea of representing each bag in the space of distribution using the principle of maximum entropy. This approach summarizes the high volume data in multi-instance data capturing the statistical properties of each bag using the $m$-dimensional sufficient statistics $E_{\hat{p}}[\phi]$. The computational complexity reduces significantly from quadratic to linear in number of instances inside each bag. We evaluated the performance of the CME in terms of rank recovery in the parameter space using the phase diagram analysis and showed accuracy of this approach in exact rank recovery (true number of parameters to model distributions). A convenient optimization framework was proposed using proximal gradient approach which is efficiently scaled to a large dataset. We used the advantage of the Lipschitz boundedness of the gradient of the maximum likelihood in the maximum entropy framework in developing the optimization algorithm.

As future research direction, one can consider the following. The rank function in CME was heuristically replaced with nuclear norm in CMEN. Nuclear norm minimization produces a low-rank solution in practice, but a theoretical characterization of when CMEN can produce the minimum rank solution was not investigated. The mathematical characterization of minimum rank solution was provided in the case where the constraints were affine \cite{recht706guaranteed}. The extension of theoretical guarantees
to the nonlinear set of inequalities is an open research direction. Our approach is an unsupervised technique in dimension reduction. Developing a new model
which accounts for the useful discriminative information in the dataset is another future research direction.

\appendices
\section{Proof of probability bound for $\sum_{i=1}^Nn_iD(p_{\hat{\lambda_i}}\|p_{\lambda_i})$}\label{appen:1}
To bound  $\sum_{i=1}^Nn_iD(p_{\hat{\lambda_i}}\|p_{\lambda_i})$, we make the following assumptions.
\paragraph{Assumption~1} \label{assump1}
$Z(\lambda)$ is strongly convex. There exists a constant $c>0$ such that
\begin{eqnarray*}
Z(\xi)  \ge  Z(\lambda) + (\xi-\lambda)^T \dot{Z}(\lambda) + \frac{c}{2} \| \xi - \lambda \|^2,
\end{eqnarray*}
for any $\xi$ and $\lambda$. The strong convexity of $Z(\lambda)$ corresponds to the fact that $\textrm{Cov}[\phi]\succeq cI$. In other words, the smallest eigenvalue of the Hessian of $Z(\lambda)$ is uniformly lower bounded everywhere by $c$. If the sufficient statistics $\phi$ is minimal (linearly independent) and $\lambda$ indexes a regular exponential family, then $Z(\lambda)$ is strictly convex \cite{amari1982differential}. The constant $c$ depends on both the sufficient statistics $\phi$ and the class of the distributions.
\paragraph{Assumption~2} \label{assump2}
$\ddot{Z}(\lambda)$ is Lipschitz continuous with constant $C$, i.e.,
\begin{eqnarray*}
&&|Z(\xi)  -  (Z(\lambda) + (\xi-\lambda)^T \dot{Z}(\lambda) + \frac{1}{2}
(\xi-\lambda)^T \ddot{Z}(\lambda) (\xi-\lambda))|\nonumber\\
&&\le  \frac{C}{6} \| \xi - \lambda \|^3,
\end{eqnarray*}
for any $\xi$ and $\lambda$. We can prove this assumption holds in our problem given $\|\phi\|_\infty \le 1$. Let $\dddot{Z}_{klm}=\frac{d^3 Z}{d\lambda_k d\lambda_l d\lambda_m}$, then $\dddot{Z}_{klm}=E_{p_{\lambda}}[\phi_k\phi_l\phi_m]-E_{p_{\lambda}}[\phi_k\phi_l]E_{p_{\lambda}}[\phi_m]-E_{p_{\lambda}}[\phi_k\phi_m]E_{p_{\lambda}}[\phi_l]-E_{p_{\lambda}}[\phi_l\phi_m]E_{p_{\lambda}}[\phi_k]+E_{p_{\lambda}}[\phi_k]E_{p_{\lambda}}[\phi_m]E_{p_{\lambda}}[\phi_l]+E_{p_{\lambda}}[\phi_m]E_{p_{\lambda}}[\phi_l]E_{p_{\lambda}}[\phi_k]$. Since each $\phi$ is bounded by $1$, therefore $\|\dddot{Z}_{klm}\|_1\le 6$.

Let $\hat\lambda = \hat\lambda_{ML}=\arg\min Z(\xi)-\xi^TE_{\hat{p}}[\phi]$ be the ML estimator of $\lambda$ as in (\ref{eq:MLEstimator}), where $E_{\hat{p}}[\phi]=\bar\phi$. The KL-divergence between $p_{\hat\lambda}=\exp{\left(\hat\lambda^T\phi-Z(\hat\lambda)\right)}$ and $p_{\lambda}=\exp{\left(\lambda^T\phi-Z(\lambda)\right)}$ is given by:
\[D(p_{\hat{\lambda}}\|p_{\lambda}) = E_{p_{\hat\lambda}}\left[\log\frac{p_{\hat\lambda}}{p_\lambda}\right]=(\hat{\lambda}-\lambda)^T \bar{\phi} - (Z(\hat{\lambda})-Z(\lambda)) \]

Note that $E_{p_{\hat\lambda}}[\phi]=E_{\hat{p}}[\phi]=\bar\phi$. Consequently, we can relate $D(p_{\hat{\lambda}}\|p_{\lambda})$ to $\delta=\bar\phi-E_{p_{\lambda}}[\phi]$ using
\begin{eqnarray}
{D(p_{\hat{\lambda}}\|p_{\lambda})=- ( \min_{\xi} D(p_{\lambda}\|p_{\xi})
- (\xi -\lambda )^T \delta )}
\label{eq:KLRelationDelta}
\end{eqnarray}
\begin{proof}
Using the definition of $\hat{\lambda}$:
\begin{eqnarray*}
D(p_{\hat{\lambda}}\|p_{\lambda}) & = & \hat{\lambda}^T E_{\hat{p}}[\phi] - Z(\hat{\lambda}) + Z(\lambda)-\lambda^T E_{\hat{p}}[\phi] \\
& = & \max_{\xi}  (\xi-\lambda)^T E_{\hat{p}}[\phi]-(Z(\xi) - Z(\lambda)) \\
& = & \max_{\xi}  (\xi-\lambda)^T E_{p_{\hat{\lambda}}}[\phi]-(Z(\xi) - Z(\lambda)) \\
& = & \max_{\xi}  (\xi-\lambda)^T E_{p_\lambda}[\phi]-(Z(\xi) - Z(\lambda))\\
& + & (\xi-\lambda)^T (E_{p_{\hat{\lambda}}}[\phi]-E_{p_\lambda}[\phi]) \\
& = & \max_{\xi}  (\xi-\lambda)^T E_{p_\lambda}[\phi]-(Z(\xi) - Z(\lambda))\\
& + & (\xi-\lambda)^T (E_{p_{\hat{\lambda}}}[\phi] -E_{p_\lambda}[\phi]) \\
& = & \max_{\xi}  -D(p_{\lambda}\|p_{\xi})
+ (\xi-\lambda)^T (E_{p_{\hat{\lambda}}}[\phi] -E_{p_\lambda}[\phi]) \\
& = &   - ( \min_{\xi} D(p_{\lambda}\|p_{\xi})- (\xi -\lambda )^T \delta ).
\end{eqnarray*}
\end{proof}
\begin{theorem}
Let $\hat{\lambda}=arg\min_{\lambda} nD(\hat{p}\|p_{\lambda})$ defined in (\ref{eq:MLEstimator}). Given assumptions (\ref{assump1}) and (\ref{assump2}), with probability at least $1-\omega_a$:
\begin{eqnarray*}
p\biggl( \sum_{i=1}^Nn_iD(p_{\hat{\lambda}_i}\|p_{\lambda_i})\ge \epsilon(\omega_a)\biggl) \le \frac{1}{a}.
\end{eqnarray*}
\end{theorem}
where $\epsilon(\omega_a)=\frac{aNm}{2}$ is an in-probability bound for the estimation error. $N$ is total number of bags and $m$ is total number of feature functions $\phi$.

To find the probability bound for the random quantity $\sum_{i=1}^Nn_iD(p_{\hat{\lambda_i}}\|p_{\lambda_i})$, we use Markov's inequality. Markov's inequality for a non-negative random variable $X$ (i.e., $p(X\ge 0)=1$) and a positive scalar $a$ is defined as follows:
\begin{eqnarray*}
p(X \ge a) \le \frac{E(X)}{a}.
\end{eqnarray*}

Markov's inequality relates the probability of random variable $X$ to its expectation. Since $\sum_{i=1}^Nn_iD(p_{\hat{\lambda}_i}\|p_{\lambda_i})$ is a non-negative value, we propose the following bound for the random quantity $\sum_{i=1}^Nn_iD(p_{\hat{\lambda}_i}\|p_{\lambda_i})$
\begin{eqnarray*}
p\biggl( \sum_{i=1}^Nn_iD(p_{\hat{\lambda}_i}\|p_{\lambda_i})\ge \epsilon(\omega_a)\biggl) \le \frac{1}{a},
\end{eqnarray*}
where $\epsilon(\omega_a) = \frac{aNm}{2}$, $N$ is the number of datasets, and $m$ is the number of feature functions. To do so, we need to compute the $E\biggl[\sum_{i=1}^Nn_iD(p_{\hat{\lambda}_i}\|p_{\lambda_i})\biggl]$. For ease of notation, we drop subscript $i$ for $\lambda_i$ and $\hat\lambda_i$. While the proof is somewhat elaborated, the outline is as follows.
\begin{itemize}
	\item relate $D(p_{\hat{\lambda}}\|p_{\lambda})$ to $\delta = E_{\hat{p}}[\phi] - E_{p_\lambda}[\phi]$
	\item express $E\biggl[D(p_{\hat{\lambda}}\|p_{\lambda})\biggl]$ in terms of moments of $\delta$
\end{itemize}
In the following, we explain each parts in details.
\subsection{relation between $D(p_{\hat{\lambda}}\|p_{\lambda})$ and $\delta$}
 We first consider the quantity $D(p_{\hat{\lambda}}\|p_{\lambda})$ and expand it as follows:
\begin{eqnarray*}\label{eq:CCMaxEntap1}
D(p_{\hat{\lambda}}\|p_{\lambda}) &=& (\hat\lambda-\lambda)^TE_{\hat{p}}[\phi] - (Z(\hat\lambda)-Z(\lambda)).
\end{eqnarray*}

Recall the $Z(\lambda) = \log \int e^{\lambda^T \phi(x)}dx$ is convex in $\lambda$.
In our analysis, we make the following assumptions.
The solution to the minimization $\xi=\hat{\lambda}$ satisfies $\dot{Z}(\hat{\lambda}) - \dot{Z}(\lambda) = \delta$.
First we analyze the term $\| \xi-\lambda\|$. Using Assumption~1, we have
\begin{eqnarray*}
(\xi-\lambda)^T (\dot{Z}(\xi) -\dot{Z}(\lambda))  \ge   c \| \xi - \lambda \|^2.
\end{eqnarray*}
This is obtained by adding $Z(\xi)  \ge  Z(\lambda) + (\xi-\lambda)^T \dot{Z}(\lambda) + \frac{c}{2} \| \xi - \lambda \|^2$ and $Z(\lambda)  \ge  Z(\xi) - (\xi-\lambda)^T \dot{Z}(\xi) + \frac{c}{2} \| \xi - \lambda \|^2$. Next, using Cauchy-Schwartz inequality: $(\xi-\lambda)^T (\dot{Z}(\xi) -\dot{Z}(\lambda)) \le
\| \xi-\lambda \| \| \dot{Z}(\xi) -\dot{Z}(\lambda)\| $ and simplifying, we obtain
\begin{eqnarray}\label{eq:4:app}
 \| \xi - \lambda \| \le \frac{1}{c} \| \dot{Z}(\xi) -\dot{Z}(\lambda)\|.
\end{eqnarray}
Finally, substituting $\xi = \hat{\lambda}$ into (\ref{eq:4:app}) and using the result $\dot{Z}(\hat{\lambda}) - \dot{Z}(\lambda) = \delta$, we obtain
\begin{eqnarray}\label{eq:boundonlambda}
&&\| \hat{\lambda} - \lambda \| \le \frac{1}{c} \| \dot{Z}(\hat{\lambda}) -\dot{Z}(\lambda)\|  = \frac{1}{c} \|\delta\|\nonumber\\
&&~~~~~= \frac{1}{c} \| \frac{1}{N} \sum_{j=1}^{n_i}\phi(x_{ij}) - E_{p_{\lambda}}[\phi(x)] \|.
\end{eqnarray}
Because $p(-1\le\phi(x)\le 1) = 1$, the Hoeffding inequality is applied. Since the probability is geometric in $n$, by the Borel-Cantelli Lemma the term on the RHS of (\ref{eq:boundonlambda}) converges to zero in the strong sense. This result guarantees that $\| \hat{\lambda} - \lambda \|$ strongly convergence to $0$ as $n_i \to \infty$ consequently making $O(\| \hat{\lambda} - \lambda \|^3)$ asymptotically negligible when compared to  $O(\| \hat{\lambda} - \lambda \|^2)$ terms.

\subsection{express $E\biggl[D(p_{\hat{\lambda}}\|p_{\lambda})\biggl]$ in terms of moments of $\delta$}
Next, we exploit Assumption~2. Note that the relation of (\ref{eq:KLRelationDelta}) suggests that $D(p_{\hat\lambda}\|p_\lambda)$ is a function of the random vector $\delta = E_{\hat{p}}[\phi]-E_{p_{\lambda}}[\phi] = \frac{1}{n}\sum[\phi(x_i)-E[\phi(x_i)]]$.

We examine a tight approximation to the relation between $D(p_{\hat\lambda}\|p_\lambda)$ and $\delta$:
\begin{equation}
D(p_{\hat\lambda}\|p_\lambda)\le \frac{1}{2} \delta^T \ddot{Z}(\lambda)^{-1} \delta
 + \frac{C\|\delta\|^3}{6c^3}
\label{eq:KLDivergenceBound}
\end{equation}

\begin{proof}
Since $D(p_{\lambda}\|p_{\xi})$ is given by $Z(\xi)$ plus affine terms in $\xi$ the Lipschitz continuity of the second derivative of $Z(\xi)$  holds also for the second derivative of  $D(p_{\lambda}\|p_{\xi})$ w.r.t. $\xi$ for any $\xi$ and $\lambda$:
\begin{eqnarray*}
| D(p_{\lambda}\|p_{\xi}) - \frac{1}{2} (\xi - \lambda)^T \ddot{Z}(\lambda) (\xi - \lambda) | \le \frac{C}{6} \|
 \xi - \lambda \|^3.
\end{eqnarray*}
Using the lower bound, we have
\begin{eqnarray*}
 D(p_{\lambda}\|p_{\xi}) \ge \frac{1}{2} (\xi - \lambda)^T \ddot{Z}(\lambda) (\xi - \lambda)  - \frac{C}{6} \|
 \xi - \lambda \|^3.
\end{eqnarray*}
Using this bound we can bound $\min_{\xi} D(p_{\lambda}\|p_{\xi}) - (\xi -\lambda )^T \delta$ as follows
\begin{eqnarray}\label{eq:abc}
&&\min_{\xi} D(p_{\lambda}\|p_{\xi}) - (\xi -\lambda )^T \delta  =
 D(p_{\lambda}\|p_{\hat{\lambda}}) - (\hat{\lambda} -\lambda )^T \delta \\
 && \ge   \frac{1}{2} (\hat{\lambda} - \lambda)^T \ddot{Z}(\lambda) (\hat{\lambda} - \lambda) - (\hat{\lambda} -\lambda )^T \delta
 - \frac{C}{6} \|  \hat{\lambda} - \lambda \|^3 \nonumber\\
 && =\frac{1}{2} (\hat{\lambda} - \lambda')^T \ddot{Z}(\lambda) (\hat{\lambda} - \lambda') -
 \frac{1}{2} \delta^T \ddot{Z}(\lambda)^{-1} \delta
 - \frac{C}{6} \|  \hat{\lambda} - \lambda \|^3 ,\nonumber
 \end{eqnarray}
where $\lambda'=\lambda-\ddot{Z}^{-1} \delta$. Substituting (\ref{eq:4:app}) in (\ref{eq:abc}) yields the lower bound:
\begin{eqnarray*}
 & \ge & \frac{1}{2} (\hat{\lambda} - \lambda')^T \ddot{Z}(\lambda) (\hat{\lambda} - \lambda') -
 \frac{1}{2} \delta^T \ddot{Z}(\lambda)^{-1} \delta
 - \frac{C\|\delta\|^3}{6c^3}  \\
 & \ge &  - ( \frac{1}{2} \delta^T \ddot{Z}(\lambda)^{-1} \delta
 + \frac{C\|\delta\|^3}{6c^3} ),
\end{eqnarray*}
where the last step is since $\ddot{Z}$ is PSD. Finally, we can bound $D(p_{\hat{\lambda}}\|p_{\lambda})$ by:
\begin{eqnarray*}
D(p_{\hat{\lambda}}\|p_{\lambda}) & = & -\min_{\xi} D(p_{\lambda}\|p_{\xi}) - (\xi -\lambda )^T \delta \\
    & \le &   \frac{1}{2} \delta^T \ddot{Z}(\lambda)^{-1} \delta  + \frac{C\|\delta\|^3}{6c^3}.
\end{eqnarray*}
Hence, $D(p_{\hat{\lambda}}\|p_{\lambda})  \le    \frac{1}{2} \delta^T \ddot{Z}(\lambda)^{-1} \delta
 + \frac{n_i C\|\delta\|^3}{6c^3}$.
\end{proof}
Next, taking the expectation on both sides of (\ref{eq:KLDivergenceBound}), we obtain:
\begin{eqnarray}
E[D(p_{\hat{\lambda}}\|p_{\lambda})]  &\le&    \frac{1}{2} E[\delta^T \ddot{Z}(\lambda)^{-1} \delta]
 + \frac{C}{6c^3}E[\|\delta\|^3] \nonumber\\
&=& \frac{1}{2} \textrm{tr}\left[\ddot{Z}(\lambda)^{-1} E[\delta \delta^T]\right]
 + \frac{C}{6c^3}E[\|\delta\|^3]
\label{eq:KLDivergenceBound}
\end{eqnarray}
Note that (\ref{eq:KLDivergenceBound}) requires only the second and third moments of $\delta$. For the moments of $\delta$, we have the following result:
\[E[\delta]=0,\]
\[E[\delta\delta^T]=\textrm{Cov}[\delta]=\frac{1}{n}\ddot{Z}(\lambda)\]

\begin{proof}
The proof of $E[\delta]=0$ is as follows:
\[E[\delta]=E[\frac{1}{n}\sum(\phi(x_i)-E[\phi(x_i)])]=\frac{1}{n}\sum(E[\phi(x_i)]-E[\phi(x_i)])=0.\]
To show that $E[\delta\delta^T]=\frac{1}{n}\ddot{Z}(\lambda)$, we first show $E[\delta\delta^T]=\frac{1}{n}\textrm{Cov}[\phi]$:
\begin{eqnarray*}
E[\delta\delta^T]&=& \frac{1}{n^2}E[\sum(\phi(x_i)-E[\phi(x_i)])\sum(\phi(x_i)-E[\phi(x_i)])^T]\\
&=& \frac{1}{n^2}\sum E[\phi(x_i)\phi(x_i)^T]-E[\phi(x_i)]E[\phi(x_i)]^T \\
&=& \frac{1}{n}\textrm{Cov}[\phi]
\end{eqnarray*}
Next, we show $\ddot{Z}(\lambda) = \textrm{Cov}[\phi]$:
\begin{eqnarray*}
\ddot{Z}(\lambda) &=& \frac{\int \phi(x)\phi(x)^T e^{\lambda^T\phi(x)}dx}{(\int e^{\lambda^T\phi(x)}dx)}\\
&-&\frac{(\int \phi(x) e^{\lambda^T\phi(x)}dx)(\int \phi(x) e^{\lambda^T\phi(x)}dx)^T}{(\int e^{\lambda^T\phi(x)}dx)}\\
&=& E[\phi(x)\phi(x)^T] - E[\phi(x)]E[\phi(x)]^T\\
&=&\textrm{Cov}[\phi].
\end{eqnarray*}
\end{proof}
Moreover, using Hoeffding's inequality, we have $\|\delta\| \le \frac{\sqrt{2\log\frac{2m}{\eta}}}{\sqrt{n}}$ w.p. at least $1-\eta$ \cite{behmardi2011entropy}.

To obtain $E[\sum_{i=1}^Nn_iD(p_{\hat{\lambda}_i}\|p_{\lambda_i})]$ we can write
\begin{eqnarray*}
E\biggl[\sum_{i=1}^Nn_iD(p_{\hat{\lambda}_i}\|p_{\lambda_i})\biggl] &=& \sum_{i=1}^NE\biggl[n_iD(p_{\hat{\lambda}_i}\|p_{\lambda_i})\biggl]\nonumber\\
&\le& \sum_{i=1}^N\frac{m}{2}\nonumber\\
&=& \frac{Nm}{2}.
\end{eqnarray*}
Therefore, using the Markov's inequality with probability $\omega_a$ where $\omega_a=\frac{1}{a}$ we have
\begin{eqnarray*}
\sum_{i=1}{N}D(p_{\hat{\lambda}_i}\|p_{\lambda_i})n_i\ge \epsilon(\omega_a),
\end{eqnarray*}
where $\epsilon(\omega_a) = \frac{aNm}{2}$.
\section{Proof of Lipschitz continuity for $\nabla g(\hat\Lambda,\Lambda)$}\label{app:lipschitz1}
In this section, we want to show that $\nabla g(\hat\Lambda,\Lambda)$ is Lipschitz continuous with constant $\tau_g = Nm$ where $N$ is total number of bags and $m$ is total number of feature functions. We prove that the Hessian matrix $\nabla^2 g(\hat\Lambda,\Lambda)$ is bounded which is stronger than Lipschitz continuity of the gradient $\nabla g(\hat\Lambda,\Lambda)$. The Hessian of $g(\hat\Lambda,\Lambda)$ is equivalent to the covariance of the feature functions $\phi$. Thus,
\begin{eqnarray*}
\nabla^2 g(\hat\Lambda,\Lambda) &=& \sum_{i=1}^{N}E_{p_{\lambda_i}}(\phi\phi^T) - \biggl(E_{p_{\lambda_i}}(\phi)E_{p_{\lambda_i}}(\phi)^T\biggl)\nonumber\\
&=& \sum_{i=1}^{N}\textrm{Cov}_{p_{\lambda_i}}(\phi).
\end{eqnarray*}
We show that the covariance of $\phi$ is bounded as follows.
\begin{eqnarray}\label{eq:bounding_cov}
\max_V \frac{V^T\textrm{Cov}_{p_{\lambda_i}}(\phi)V}{V^TV} &=& \max_V\biggl( \frac{E_{p_{\lambda_i}}[(V^T\phi)^2]}{V^TV} - \frac{[E_{p_{\lambda_i}}(V^T\phi)]^2}{V^TV}\biggl)\nonumber\\
&\le& \max_V \frac{E_{p_{\lambda_i}}[(V^T\phi)^2]}{V^TV}.
\end{eqnarray}
Note that $\forall V, \frac{[E_{p_{\lambda_i}}(V^T\phi)]^2}{V^TV}\ge 0$. By Cauchy-Schwartz inequality, we have
\begin{eqnarray*}
(V^T\phi)^2 \le V^TV\phi^T\phi.
\end{eqnarray*}
Since $\phi^T\phi = \sum_{i=1}^m \phi_i^2$ and $\|\phi_i\|_\infty=1$, $\phi^T\phi \le m$. Hence,
\begin{eqnarray}\label{eq:bounding_vphi}
(V^T\phi)^2 \le V^TV m.
\end{eqnarray}
Substituting (\ref{eq:bounding_vphi}) into (\ref{eq:bounding_cov}), we obtain
\begin{eqnarray*}
\sum_{i=1}^{N}\max_V \frac{V^T\textrm{Cov}_{p_{\lambda_i}}(\phi)V}{V^TV} \le \sum_{i=1}^{N}\max_V \frac{E_{p_{\lambda_i}}[V^TVm]}{V^TV} = Nm.
\end{eqnarray*}

\ifCLASSOPTIONcaptionsoff
  \newpage
\fi



%
\bibliographystyle{IEEEbib}
\bibliography{SSPbib}

\begin{thebibliography}{10}

\bibitem{dietterich1997solving}
T.G. Dietterich, R.H. Lathrop, and T.~Lozano-P{\'e}rez,
\newblock ``Solving the multiple instance problem with axis-parallel
  rectangles,''
\newblock {\em Artificial Intelligence}, vol. 89, no. 1-2, pp. 31--71, 1997.

\bibitem{ni2008multi}
K.~Ni, J.~Paisley, L.~Carin, and D.~Dunson,
\newblock ``Multi-task learning for analyzing and sorting large databases of
  sequential data,''
\newblock {\em IEEE Transactions on Signal Processing}, vol. 56, no. 8, pp.
  3918--3931, 2008.

\bibitem{andrews2002support}
S.~Andrews, I.~Tsochantaridis, and T.~Hofmann,
\newblock ``Support vector machines for multiple-instance learning,''
\newblock {\em Proceedings of Advances in Neural Information Processing
  Systems}, vol. 15, pp. 561--568, 2002.

\bibitem{qi2007music}
Y.~Qi, J.W. Paisley, and L.~Carin,
\newblock ``Music analysis using hidden markov mixture models,''
\newblock {\em IEEE Transactions on Signal Processing}, vol. 55, no. 11, pp.
  5209--5224, 2007.

\bibitem{viola2006multiple}
P.~Viola, J.~Platt, and C.~Zhang,
\newblock ``Multiple instance boosting for object detection,''
\newblock in {\em Proceedings of Advances in Neural Information Processing
  Systems}, 2006, vol.~18, pp. 1417--1426.

\bibitem{zhang2002content}
Q.~Zhang, S.A. Goldman, W.~Yu, and J.E. Fritts,
\newblock ``Content-based image retrieval using multiple-instance learning,''
\newblock in {\em Proceedings of International Workshop on Machine Learning},
  2002, pp. 682--689.

\bibitem{Wang2000}
Jun Wang, Zucker, and Jean-Daniel,
\newblock ``Solving multiple-instance problem: A lazy learning approach,''
\newblock in {\em Proceedings of International Conference on Machine Learning},
  Pat Langley, Ed., 2000, pp. 1119--1125.

\bibitem{blei2003latent}
D.M. Blei, A.Y. Ng, and M.I. Jordan,
\newblock ``{Latent Dirichlet Allocation},''
\newblock {\em Journal of Machine Learning Research}, vol. 3, pp. 993--1022,
  2003.

\bibitem{ramon2000multi}
J.~Ramon and L.~De~Raedt,
\newblock ``Multi instance neural networks,''
\newblock in {\em Proceedings of ICML-2000, Workshop on Attribute-Value and
  Relational Learning}, 2000, pp. 53--60.

\bibitem{maron1998framework}
O.~Maron and T.~Lozano-P{\'e}rez,
\newblock ``A framework for multiple-instance learning,''
\newblock in {\em Proceedings of Advances in Neural Information Processing
  Systems}, 1998, pp. 570--576.

\bibitem{zhang2001dd}
Q.~Zhang and S.A. Goldman,
\newblock ``Em-dd: An improved multiple-instance learning technique,''
\newblock in {\em Proceedings of Advances in Neural Information Processing
  Systems}. 2001, vol.~14, pp. 1073--1080, Cambridge, MA: MIT Press.

\bibitem{xu2004logistic}
X.~Xu and E.~Frank,
\newblock ``Logistic regression and boosting for labeled bags of instances,''
\newblock {\em Advances in Knowledge Discovery and Data Mining}, pp. 272--281,
  2004.

\bibitem{zhang2009multi}
M.L. Zhang and Z.H. Zhou,
\newblock ``Multi-instance clustering with applications to multi-instance
  prediction,''
\newblock {\em Applied Intelligence}, vol. 31, no. 1, pp. 47--68, 2009.

\bibitem{xu2011multi}
Y.~Xu, W.~Ping, and A.T. Campbell,
\newblock ``Multi-instance metric learning,''
\newblock in {\em Proceedings of IEEE International Conference on Data Mining},
  2011, pp. 874--883.

\bibitem{gärtner2002multi}
T.~G{\"a}rtner, P.A. Flach, A.~Kowalczyk, and A.J. Smola,
\newblock ``Multi-instance kernels,''
\newblock in {\em Proceedings of International Conference on Machine Learning},
  2002, pp. 179--186.

\bibitem{maron1998multiple}
O.~Maron and A.L. Ratan,
\newblock ``Multiple-instance learning for natural scene classification,''
\newblock in {\em Proceedings of International Conference on Machine Learning},
  1998, vol.~15, pp. 341--349.

\bibitem{paisley2010active}
J.~Paisley, X.~Liao, and L.~Carin,
\newblock ``Active learning and basis selection for kernel-based linear models:
  a bayesian perspective,''
\newblock {\em IEEE Transactions on Signal Processing}, vol. 58, no. 5, pp.
  2686--2700, 2010.

\bibitem{amari1982differential}
S.I. Amari,
\newblock ``Differential geometry of curved exponential families-curvatures and
  information loss,''
\newblock {\em The Annals of Statistics}, pp. 357--385, 1982.

\bibitem{csiszar2004information}
I.~Csisz{\'a}r and P.C. Shields,
\newblock {\em Information theory and statistics: A tutorial}, vol.~1,
\newblock Communication and information theroy, 2004.

\bibitem{behmardi2011entropy}
B.~Behmardi, R.~Raich, and A.O. Hero,
\newblock ``Entropy estimation using the principle of maximum entropy,''
\newblock in {\em Proceedings of IEEE International Conference on Acoustics,
  Speech and Signal Processing}. IEEE, 2011, pp. 2008--2011.

\bibitem{jaynes1957information}
E.T. Jaynes,
\newblock ``Information theory and statistical mechanics,''
\newblock {\em Physical review}, vol. 106, no. 4, pp. 620, 1957.

\bibitem{berger1996maximum}
A.L. Berger, V.J.D. Pietra, and S.A.D. Pietra,
\newblock ``A maximum entropy approach to natural language processing,''
\newblock {\em Computational linguistics}, vol. 22, no. 1, pp. 39--71, 1996.

\bibitem{dudik2007maximum}
M.~Dud{\i}k, S.J. Phillips, and R.E. Schapire,
\newblock ``Maximum entropy density estimation with generalized regularization
  and an application to species distribution modeling,''
\newblock {\em Journal of Machine Learning Research}, vol. 8, pp. 1217--1260,
  2007.

\bibitem{zhu2005multi}
S.~Zhu, X.~Ji, W.~Xu, and Y.~Gong,
\newblock ``Multi-labelled classification using maximum entropy method,''
\newblock in {\em Proceedings of the 28th annual international ACM SIGIR
  conference on Research and development in information retrieval}. ACM, 2005,
  pp. 274--281.

\bibitem{skilling1984maximum}
J.~Skilling and RK~Bryan,
\newblock ``Maximum entropy image reconstruction-general algorithm,''
\newblock {\em Monthly Notices of the Royal Astronomical Society}, vol. 211,
  pp. 111, 1984.

\bibitem{boyd2004convex}
S.P. Boyd and L.~Vandenberghe,
\newblock {\em Convex optimization},
\newblock Cambridge University Press, 2004.

\bibitem{della1997inducing}
S.~Della~Pietra, V.~Della~Pietra, and J.~Lafferty,
\newblock ``Inducing features of random fields,''
\newblock {\em IEEE Transactions on Pattern Analysis and Machine Intelligence},
  vol. 19, no. 4, pp. 380--393, 1997.

\bibitem{salakhutdinov2003convergence}
R.~Salakhutdinov, S.T. Roweis, Z.~Ghahramani, et~al.,
\newblock ``On the convergence of bound optimization algorithms,''
\newblock in {\em Uncertainty in Artificial Intelligence}, 2003, vol.~19, pp.
  509--516.

\bibitem{krishnapuram2005sparse}
B.~Krishnapuram, L.~Carin, M.A.T. Figueiredo, and A.J. Hartemink,
\newblock ``Sparse multinomial logistic regression: Fast algorithms and
  generalization bounds,''
\newblock {\em IEEE Transactions on Pattern Analysis and Machine Intelligence},
  vol. 27, no. 6, pp. 957--968, 2005.

\bibitem{dudik2004performance}
M.~Dudik, S.~Phillips, and R.~Schapire,
\newblock ``Performance guarantees for regularized maximum entropy density
  estimation,''
\newblock {\em Learning Theory}, pp. 472--486, 2004.

\bibitem{chen2000survey}
S.F. Chen and R.~Rosenfeld,
\newblock ``A survey of smoothing techniques for me models,''
\newblock {\em IEEE Transactions on Speech and Audio Processing}, vol. 8, no.
  1, pp. 37--50, 2000.

\bibitem{dudik2007hierarchical}
M.~Dudik, D.M. Blei, and R.E. Schapire,
\newblock ``Hierarchical maximum entropy density estimation,''
\newblock in {\em Proceedings of the 24th international conference on Machine
  learning}. ACM, 2007, pp. 249--256.

\bibitem{recht706guaranteed}
B.~Recht, M.~Fazel, and P.A. Parrilo,
\newblock ``{Guaranteed minimum-rank solutions of linear matrix equations via
  nuclear norm minimization, 2007},''
\newblock {\em SIAM Review}, vol. 52, pp. 471--501, 2010.

\bibitem{candes2010matrix}
E.J. Candes and Y.~Plan,
\newblock ``{Matrix completion with noise},''
\newblock {\em Proceedings of the IEEE}, vol. 98, no. 6, pp. 925--936, 2010.

\bibitem{srebro2005maximum}
N.~Srebro, J.D.M. Rennie, and T.~Jaakkola,
\newblock ``Maximum-margin matrix factorization,''
\newblock in {\em Proceedings of Conference on Advances in Neural Information
  Processing Systems}, 2005, vol.~17, pp. 1329--1336.

\bibitem{pong2009trace}
T.K. Pong, P.~Tseng, S.~Ji, and J.~Ye,
\newblock ``Trace norm regularization: Reformulations, algorithms, and
  multi-task learning,''
\newblock {\em Submitted to SIAM Journal on Optimization}, 2009.

\bibitem{behmardissp2}
B.~Behmardi, F.~Briggs, X.~Fern, and R.~Raich,
\newblock ``Regularized joint density estimation for multi-instance learning,''
\newblock in {\em Proceedings of IEEE International Workshop on Statistical
  Signal Processing}, 2012, pp. 740--743.

\bibitem{gill1981practical}
P.E. Gill, W.~Murray, and M.H. Wright,
\newblock {\em Practical optimization}, vol.~1,
\newblock Academic press, 1981.

\bibitem{behmardiTSP}
B.~Behmardi and R.~Raich,
\newblock ``On confindence-constrained rank recovery in topic models,''
\newblock {\em IEEE Transactions on Signal Processing}, vol. 60, no. 10, pp.
  5146--5162, 2012.

\bibitem{candes2009robust}
E.J. Candes, X.~Li, Y.~Ma, and J.~Wright,
\newblock ``{Robust principal component analysis},''
\newblock {\em Journal of ACM}, vol. 58, no. 1, pp. 1--37, 2009.

\bibitem{meka2008rank}
R.~Meka, P.~Jain, C.~Caramanis, and I.S. Dhillon,
\newblock ``Rank minimization via online learning,''
\newblock in {\em Proceedings of the 25th International Conference on Machine
  learning}. ACM, 2008, pp. 656--663.

\bibitem{haldar2009rank}
J.P. Haldar and D.~Hernando,
\newblock ``Rank-constrained solutions to linear matrix equations using
  powerfactorization,''
\newblock {\em Signal Processing Letters}, vol. 16, no. 7, pp. 584--587, 2009.

\bibitem{liu2009implementable}
Y.J. Liu, D.~Sun, and K.C. Toh,
\newblock ``{An implementable proximal point algorithmic framework for nuclear
  norm minimization},''
\newblock {\em Mathematical Programming}, pp. 1--38, 2009.

\bibitem{toh2010accelerated}
K.C. Toh and S.~Yun,
\newblock ``{An accelerated proximal gradient algorithm for nuclear norm
  regularized linear least squares problems},''
\newblock {\em Pacific Journal of Optimization}, vol. 6, pp. 615--640, 2010.

\bibitem{cai810singular}
J.F. Cai, E.J. Candes, and Z.~Shen,
\newblock ``{A singular value thresholding algorithm for matrix completion},''
\newblock {\em Journal on Optimization}, vol. 20, pp. 615--640, 2008.

\bibitem{nesterov1983method}
Y.~Nesterov,
\newblock ``{A method of solving a convex programming problem with convergence
  rate O (1/k2)},''
\newblock {\em Soviet Mathematics Doklady}, vol. 27, pp. 372--376, 1983.

\bibitem{lin2009augmented}
Z.~Lin, M.~Chen, L.~Wu, and Y.~Ma,
\newblock ``{The augmented {L}agrange multiplier method for exact recovery of
  corrupted low-rank matrices},''
\newblock {\em Mathematical Programming}, 2009.

\bibitem{donoho2006sparse}
D.L. Donoho, I.~Drori, Y.~Tsaig, and J.L. Starck,
\newblock ``{Sparse solution of underdetermined linear equations by stagewise
  orthogonal matching pursuit},''
\newblock {\em Citeseer}, 2006.

\bibitem{rahimi2007random}
A.~Rahimi and B.~Recht,
\newblock ``Random features for large-scale kernel machines,''
\newblock {\em Proceedings of Advances in Neural Information Processing
  Systems}, vol. 20, pp. 1177--1184, 2007.

\bibitem{ma2011fixed}
S.~Ma, D.~Goldfarb, and L.~Chen,
\newblock ``Fixed point and bregman iterative methods for matrix rank
  minimization,''
\newblock {\em Mathematical Programming}, vol. 128, no. 1, pp. 321--353, 2011.

\bibitem{efron2004least}
B.~Efron, T.~Hastie, I.~Johnstone, and R.~Tibshirani,
\newblock ``Least angle regression,''
\newblock {\em The Annals of statistics}, vol. 32, no. 2, pp. 407--499, 2004.

\bibitem{hale2007fixed}
E.T. Hale, W.~Yin, and Y.~Zhang,
\newblock ``A fixed-point continuation method for l1-regularized minimization
  with applications to compressed sensing,''
\newblock {\em CAAM TR07-07, Rice University}, 2007.

\bibitem{stewart1993early}
G.W. Stewart,
\newblock ``On the early history of the singular value decomposition,''
\newblock {\em SIAM review}, vol. 35, no. 4, pp. 551--566, 1993.

\bibitem{duygulu2006object}
P.~Duygulu, K.~Barnard, J.~De~Freitas, and D.~Forsyth,
\newblock ``{Object recognition as machine translation: Learning a lexicon for
  a fixed image vocabulary},''
\newblock {\em Proceedigs of European Conference on Computer Vision}, pp.
  349--354, 2006.

\bibitem{carter2009information}
K.M. Carter, R.~Raich, W.G. Finn, and A.O. Hero,
\newblock ``Information preserving component analysis: Data projections for
  flow cytometry analysis,''
\newblock {\em Selected Topics in Signal Processing, IEEE Journal of}, vol. 3,
  no. 1, pp. 148--158, 2009.

\bibitem{terrell1990maximal}
George~R Terrell,
\newblock ``The maximal smoothing principle in density estimation,''
\newblock {\em Journal of the American Statistical Association}, vol. 85, no.
  410, pp. 470--477, 1990.

\end{thebibliography}

\end{document}